\documentclass[12pt]{article}

\usepackage[utf8]{inputenc}
\usepackage[T1]{fontenc}
\usepackage{amsmath,amssymb,amsthm,mathtools}
\usepackage{booktabs,tabularx,array,longtable}
\usepackage[margin=1in]{geometry}
\usepackage[round]{natbib}
\usepackage{subcaption}
\usepackage[colorlinks=true,linkcolor=blue,citecolor=blue,urlcolor=blue]{hyperref}
\usepackage{cleveref}

\usepackage{setspace}
\usepackage{enumitem}
\usepackage{graphicx}
\graphicspath{{figures/}}
\usepackage{placeins}
\usepackage{xurl}
\newtheorem{theorem}{Theorem}
\newtheorem{proposition}{Proposition}

\newtheorem{lemma}{Lemma}

\theoremstyle{definition}
\newtheorem{definition}{Definition}
\crefname{assumption}{Assumption}{Assumptions}
\Crefname{assumption}{Assumption}{Assumptions}

\newcommand{\E}{\mathbb{E}}

\newcommand{\KL}{D_{\mathrm{KL}}}
\newcommand{\Rcal}{\mathcal{R}}
\newcommand{\mD}{\mathrm{D}}
\newcommand{\mL}{\mathrm{L}}
\newcommand{\mA}{\mathrm{A}}
\newcommand{\mF}{\tilde{\mathrm{A}}}
\newcommand{\mM}{\mathrm{M}}
\newcommand{\kD}{\kappa^{\mD}}
\newcommand{\kL}{\kappa^{\mL}}
\newcommand{\kA}{\kappa^{\mA}}
\newcommand{\fbar}{\bar f}

\newcounter{mechanism}
\makeatletter
\newenvironment{mechanism}[3]{%
  \begin{figure}[htbp]\noindent
  \refstepcounter{mechanism}\def\@currentlabel{#1}\label{#3}%
  \begin{minipage}{\textwidth}%
  \hrule height 0.8pt\relax\vspace{1.2ex}%
  \noindent\textbf{Mechanism #1: #2}\par\vspace{0.9ex}%
  \hrule\relax
  \begin{description}[leftmargin=0pt,labelindent=0pt,itemsep=0.4ex,topsep=0.8ex,font=\normalfont\itshape]%
}{%
  \end{description}\vspace{0.6ex}%
  \hrule height 0.8pt\relax
  \end{minipage}\end{figure}}
\makeatother

\begin{document}
\pagenumbering{roman}

\title{%
{~~\\[-1in]
\Huge\bf Adapting the Process of Adaptation:\\[.06in]
Darwinian, Lamarckian, and \\[.125in]
Agent-Directed Evolution in AI\thanks{Portions of this manuscript were presented at the MIT CSAIL 2026 Agentic AI Summmit, the 6th Research in Behavioral Finance Conference in Amsterdam (2026), and the RESI-MIT seminar, and we thank participants in those forums for useful feedback. The views expressed in this article are those of the authors alone and do not necessarily reflect the views of any institution or agency with which they are affiliated, or any of the individuals  acknowledged above. AI was used in the preparation of this manuscript (see Appendix \ref{app:ai_use}).}}}

 \author{Andrew W.~Lo\thanks{MIT Sloan School of Management; MIT Laboratory for Financial Engineering; MIT Computer Science
 and Artificial Intelligence Laboratory; Santa Fe Institute; $100$ Main Street, E$62$--$618$, Cambridge, MA $02142$, \url{alo-admin@mit.edu}.}
 \ and Chaoyi Zhao\thanks{MIT Sloan School of Management; MIT Laboratory for Financial Engineering. $1$ Main Street, E$90$, Cambridge, MA $02142$, \url{cy_zhao@mit.edu}.}}

\date{This Draft: 20 November 2026}

\maketitle \thispagestyle{empty}
\centerline{\bf Abstract}
\vskip 10pt
\noindent%
{Evolution has shaped intelligence for billions of years. Artificial intelligence (AI) now raises a striking possibility: intelligence may begin to direct its own evolution. As AI agents become capable of modifying and evaluating their own behavior, how quickly could they improve, and how can that improvement remain safe? We develop a theoretical framework comparing Darwinian, Lamarckian, and agent-directed evolution following environmental change. Darwinian evolution relies on natural selection, Lamarckian evolution adds inherited learning, and agent-directed evolution allows agents to generate and evaluate behavioral variants and choose how extensively to search. When learning is sufficiently effective, these mechanisms exhibit progressively faster adaptation. Generating more variants accelerates adaptation, while optimal search effort depends on learning ability, problem difficulty, and experimentation costs. However, an agent's ability to direct its improvement does not ensure that improvements are evaluated against the intended objective. When the evaluation target is misspecified, additional search accelerates convergence toward the wrong behavior, even as measured performance improves. In recursive self-improvement, such errors may persist across generations, posing challenges for AI safety. Independent evaluation and oversight must keep pace with autonomous AI development. The central challenge is not simply to accelerate the evolution of intelligence, but to preserve humanity's ability to guide its direction, limit its speed, and intervene when necessary.}
\baselineskip 14pt
\vskip 10pt \noindent


\newpage
\thispagestyle{empty}
{\singlespacing\small\setlength{\parskip}{0pt}\setlength{\baselineskip}{11pt}\tableofcontents}
\newpage
\pagenumbering{arabic}
\setlength{\baselineskip}{1.5\baselineskip}
\onehalfspacing

\section{Introduction}\label{sec:intro}

On 8 September 2026, OpenAI announced that a coordinated system of artificial intelligence (AI) agents had produced a proposed resolution of the Navier--Stokes Millennium Prize problem, showing that a finite-time singularity forms under smooth external forcing \citep{openai2026navier}. According to OpenAI, roughly 10,000 concurrent agents explored different formulations and approaches, shared intermediate findings, and built on promising results. They reached the proposed solution in about 88 hours and then spent 17 hours formalizing and verifying it in Lean \citep{demoura2021lean}.

This search resembles evolution: variants are generated, promising ones are retained, and useful information is passed on to later attempts. FunSearch \citep{romeraparedes2024funsearch}, AlphaEvolve \citep{novikov2025alphaevolve}, and the Darwin G\"odel Machine \citep{zhang2025darwin} follow the same principle: AI systems propose modifications to programs or agents, evaluate them, and build on the successful ones.

These ideas come together in an emerging research direction, \emph{recursive self-\hspace{0pt}improvement} (RSI), in which an AI system improves itself and uses the improved version to make further improvements. In the Darwin G\"odel Machine, AI agents modify their own code, evaluate the resulting variants, and keep the successful changes for later generations \citep{zhang2025darwin}. More recently, \citet{srikanth2026recursive} developed AIDE$^2$, an AI research agent that recursively modifies its own code and evaluates the results. These studies suggest that AI systems are increasingly able not only to perform complex tasks but also to improve the processes through which they perform them.

These developments point to a more fundamental possibility: AI systems may increasingly take control of their own evolution. Instead of improving within a process designed by humans, a system may decide how it searches for improvements, how extensively it experiments, and how it carries improvements forward to later generations. Because improved systems can produce further improvements, this could greatly accelerate AI development and make its capabilities harder for humans to anticipate and oversee. In an early conceptual essay, \citet{farmer1991artificial} anticipated this possibility, arguing that control over the evolution of artificial organisms could pass from humans to the organisms themselves, and that evolution would then be far faster than biological evolution, because acquired characteristics would be transmitted directly and innovation would come from design rather than random variation. We call this shift \emph{agent-directed evolution}. It holds great potential for scientific discovery and technological progress, but it also raises safety concerns as AI systems gain greater control over their own development.

This possibility motivates several questions. How much faster can a system adapt through learning and self-directed search than through natural selection alone? What determines the speed of adaptation, and how should an agent weigh the benefits of more search against its costs? More importantly, does faster adaptation necessarily lead to better outcomes, or can it also accelerate the development of behaviors that humans did not intend?

To address these questions, we build on the evolutionary framework of \citet{brennan2011origin,brennan2012evolutionary}, in which individuals inherit different behaviors and natural selection favors the behaviors that lead to faster population growth. When the environment changes, previously successful behaviors may no longer be optimal, and the population must adapt. In \Cref{sec:model}, we show that the growth lost to suboptimal behavior equals the Kullback--Leibler (KL) divergence between the environmental and behavioral distributions (\Cref{prop:kl,prop:average}). We call this loss \emph{growth regret} and use it to compare how quickly Darwinian, Lamarckian, and agent-directed evolution adapt. This comparison gives a quantitative form to the acceleration that \citet{farmer1991artificial} anticipated over three decades ago.

The three mechanisms start from the same population. Under \emph{Darwinian evolution}, inherited behaviors are fixed, and the population adapts as natural selection makes better-adapted individuals more prevalent (\Cref{sec:darwinian}). Under \emph{Lamarckian evolution}, individuals can also learn from their environment and pass their acquired improvements to later generations (\Cref{sec:lamarckian}). Under \emph{agent-directed evolution}, individuals further generate multiple variants of themselves, evaluate them, and retain the best for further learning (\Cref{sec:agentic}). More importantly, the agent decides how extensively to search, which we use as a simple model of how an intelligent system can influence its own evolution. We measure speed by the \emph{adaptation time}, the number of generations needed to reduce growth regret to a given threshold, and the \emph{adaptation rate}, the long-run exponential rate at which growth regret declines.

Our first set of results characterizes the adaptation speeds of the three mechanisms. Darwinian adaptation relies on natural selection to spread individuals who already have the optimal behavior, so it can take arbitrarily long when they are initially rare (\Cref{thm:darwin}). Lamarckian evolution removes this dependence, because individuals improve their inherited behavior directly through learning (\Cref{thm:lam}). Agent-directed evolution adapts even faster by searching among alternative improvements (\Cref{thm:agent}). In \Cref{sec:combined}, we show that, when learning is sufficiently effective, agent-directed evolution has the highest long-run adaptation rate, followed by Lamarckian and then Darwinian evolution; the same ranking holds for expected adaptation times (\Cref{thm:hierarchy}).

Our second set of results, in \Cref{sec:agentic}, concerns how search accelerates adaptation and how an agent should allocate its search effort. Generating more variants always increases the adaptation rate, by an amount that depends on the difficulty of the problem (\Cref{prop:gain}). For easier problems, the gain can be substantial: in the most favorable case, it grows approximately as $2\log n$, where $n$ is the number of variants. For harder problems, the gain is smaller and approaches a finite ceiling, although more variants still speed up adaptation. When search is costly, the optimal number of variants increases as learning becomes less effective, the problem becomes easier, or search becomes cheaper (\Cref{prop:Nstar}). An intelligent system can thus adjust its search effort according to its learning ability, problem difficulty, and available resources.

Our third set of results concerns AI safety: faster adaptation does not necessarily move a system toward the objectives its developers intend. In \Cref{sec:safety}, we extend the agent-directed mechanism so that search and learning are guided by an incorrectly specified evaluation target. The agent then keeps improving rapidly according to its evaluator, but its true growth regret does not vanish, and more variants only accelerate convergence toward the incorrect target (\Cref{thm:pressure}). Natural selection, acting on realized outcomes, can eventually correct the population's behavior, but this correction may be slow. Independent evaluation and oversight therefore become increasingly important.

Our analysis brings together ideas from evolutionary biology, economics, and AI: drawing on the biological mechanisms of selection and inheritance, we use a growth-based framework to study how learning and search affect the speed and cost of adaptation in AI systems. Appendix~\ref{app:notation} summarizes how the main elements of our model correspond to their counterparts in evolutionary biology and modern AI systems.

The remainder of this paper proceeds as follows. \Cref{sec:lit} reviews the related literature, and \Cref{sec:model} introduces the Brennan--Lo framework. \Cref{sec:darwinian,sec:lamarckian,sec:agentic} analyze Darwinian, Lamarckian, and agent-directed evolution, respectively, and \Cref{sec:combined} compares their adaptation speeds. \Cref{sec:safety} examines adaptation toward an incorrectly specified target and its implications for AI safety. \Cref{sec:conclusion} concludes.

\section{Related Literature}\label{sec:lit}

We review six areas of research in evolutionary biology, information theory, economics, and AI that bear on the speed and direction of adaptation.

\paragraph{Natural Selection and Growth-Optimal Behavior.}
We build on the evolutionary model of \citet{brennan2011origin}, in which natural selection favors inherited behaviors that lead to higher population growth. Later studies extend this framework to bounded rationality and intelligence \citep{brennan2012evolutionary}, behavioral diversification \citep{brennan2018variety}, and learning under limited information and memory \citep{lozhang2021bayesian,lo2025discrimination}. Two companion papers apply this framework to AI. \citet{brennan2026socialization} study how selection and learning shape an AI agent's choice between completing a task and observing a rule of conduct. \citet{lo2026coevolution} study when natural selection in interacting human and artificial populations produces artificial behavior that benefits humans and whether that behavior lasts. In evolutionary biology, \emph{bet hedging} describes how organisms diversify their offspring's behaviors to cope with environmental uncertainty \citep{cohen1966optimizing,bull1987evolution,seger1987bethedging}. Building on this literature, we study how quickly a population adapts after an environmental change. 

\paragraph{Information Theory and Evolutionary Growth.}
Information and growth have long been linked. In the framework of \citet{kelly1956}, the growth lost by using incorrect probabilities can be expressed in terms of the KL divergence \citep[Chapter~6]{cover2006elements}, and evolutionary biology studies the fitness value of information about the environment \citep{kussell2005phenotypic,donaldsonmatasci2010fitness,rivoire2011value}. We establish an analogous relationship between population growth and KL divergence, which lets us measure the loss from suboptimal behavior and compare how quickly different evolutionary mechanisms eliminate it.

\paragraph{Learning and the Inheritance of Acquired Traits.}
Previous research distinguishes learning that affects an individual's performance from learning whose improvements are inherited. In the Baldwin effect, individual learning can facilitate evolution even without the inheritance of acquired traits \citep{hinton1987learning,ackley1991interactions}, whereas Lamarckian evolutionary algorithms pass improvements acquired through learning to offspring \citep{whitley1994lamarckian}. Related ideas appear in population-based training \citep{jaderberg2017pbt} and in methods that transfer learned knowledge across generations of neural networks \citep{feng2025learngene}. Our Lamarckian mechanism isolates the inheritance of acquired improvements and examines how this channel changes the speed of adaptation under natural selection.

\paragraph{Adaptive Search and Resource Allocation.}
Evolutionary computation has long studied how to adjust search parameters, such as mutation rates and the distribution of candidate solutions, to improve performance \citep{eiben1999parameter,hansen2001completely,bedau2003evolvability,ollivier2017information}; autoconstructive evolution goes further and lets the mechanisms of reproduction and variation evolve within the individuals themselves \citep{spector2001autoconstructive}. Economic models examine how agents adjust their learning rules to changing environments or forecasting performance \citep{evans2001learning,marcet2003recurrent}. More recent research on inference-time computation studies how the benefit of generating more candidates depends on problem difficulty and computational resources \citep{brown2024monkeys,snell2024scaling}. Our agent-directed mechanism poses a related decision in an evolutionary setting, how many variants to generate, and we characterize how search raises the rate of adaptation and how the optimal number of variants depends on problem difficulty, learning ability, and search costs.

\paragraph{RSI in AI.}
The possibility that intelligent machines could improve their own capabilities has been discussed since \citet{good1965speculations}; an early formal approach is the G\"odel machine of \citet{schmidhuber2007godel}. FunSearch and AlphaEvolve use AI to generate, evaluate, and retain improved solutions \citep{romeraparedes2024funsearch,novikov2025alphaevolve}, and the Darwin G\"odel Machine \citep{zhang2025darwin}, AIDE$^2$ \citep{srikanth2026recursive}, and multi-agent RSI \citep{lee2026recursive} extend this process to AI agents themselves. These developments have prompted discussion of whether AI evolution will remain directed by humans or become increasingly autonomous \citep{boudry2026domesticated,muller2026evolvable,zong2026coevolution}. Our analysis offers a theoretical perspective by studying how an agent's control over its own search effort affects the speed and cost of adaptation.

\paragraph{Evaluation and the Direction of Adaptation.}
The effectiveness of search depends on how candidate improvements are evaluated. Evolutionary algorithms can exploit evaluation criteria in unintended ways \citep{lehman2020surprising}, and stronger optimization against learned reward models can widen the gap between measured reward and true quality \citep{gao2023scaling}. We examine how an incorrectly specified evaluation target affects evolutionary adaptation and show that more search can accelerate adaptation toward the evaluator's target without correcting the error.

\section{The Framework}\label{sec:model}

To compare different mechanisms of adaptation, we need a framework that identifies both the optimal behavior and the cost of deviating from it. We adopt the evolutionary model of \citet{brennan2011origin}, in which natural selection favors behaviors that maximize long-run population growth. \Cref{ssec:lineage} studies the growth of a lineage, \Cref{ssec:population} extends the analysis to a population with multiple lineages, and \Cref{ssec:shift} introduces the three evolutionary mechanisms, establishes a common setting for comparing them, and defines the adaptation time and rate.

\subsection{Growth of a lineage}\label{ssec:lineage}

Consider a large lineage of asexually reproducing individuals. 
In generation $t$, all individuals inherit the same behavior $f_t\in(0,1)$, choosing action $a$ with probability $f_t$ and action $b$ with probability $1-f_t$. The three evolutionary mechanisms considered in this study differ in whether and how $f_t$ changes across generations. Individuals in generation $t$ reproduce to form generation $t+1$, and their reproductive success depends on the chosen action and the environmental state $X_{t+1}\in\{\alpha,\beta\}$, which is shared by all individuals. Let $x_a$ and $x_b$ denote the numbers of offspring produced by an individual choosing $a$ and $b$, respectively. Then
\[
(x_a,x_b)=
\left\{
\begin{array}{lll}
(m,0), & X_{t+1}=\alpha, & \text{with probability }p,\\[2pt]
(0,m), & X_{t+1}=\beta, & \text{with probability }1-p,
\end{array}
\right.
\]
where $m>0$ is the number of offspring produced by an individual choosing the successful action: action $a$ succeeds in state $\alpha$, and action $b$ succeeds in state $\beta$. Individuals choosing the other action produce no offspring. The probability of state $\alpha$ is $p\in(0,1)$, and the environmental states $X_1,X_2,\ldots$ are independent across generations.

Let $N_t$ denote the size of the lineage in generation $t$. Because the lineage is large, a fraction $f_t$ of its individuals chooses action $a$, while the remaining fraction $1-f_t$ chooses action $b$. If $X_{t+1}=\alpha$, only those choosing $a$ reproduce, each producing $m$ offspring, so the lineage grows by the factor $N_{t+1}/N_t=mf_t$. Similarly, if $X_{t+1}=\beta$, the lineage grows by the factor $N_{t+1}/N_t=m(1-f_t)$. Taking the expected logarithm of these growth factors gives
\begin{equation}
\begin{aligned}
I_p(f_t)\equiv\E\Big[\log\frac{N_{t+1}}{N_t}\,\Big|\,f_t\Big]
&=p\log(mf_t)+(1-p)\log\big(m(1-f_t)\big)\\
&=\log m+p\log f_t+(1-p)\log(1-f_t).
\end{aligned}
\label{eq:growth}
\end{equation}
We refer to $I_p(f)$ as the growth rate associated with behavior $f$.

Given the growth rate in Equation~\eqref{eq:growth}, we can now identify the inherited behavior that maximizes population growth.\footnote{Proofs of the formal results are in Appendix~\ref{app:proofs}.}

\begin{proposition}[Probability matching]\label{prop:pm}
For $p\in(0,1)$, $I_p$ is strictly concave on $(0,1)$, and
\[
\arg\max_{f\in(0,1)}I_p(f)=p.
\]
\end{proposition}

The proposition shows that a lineage maximizes its expected log growth in every generation by matching the fraction of individuals choosing each action to the probability that the action succeeds. The intuition is that diversification is essential when all individuals face the same environmental risk. A lineage that always chooses one action will eventually become extinct when the other action succeeds, whereas a lineage that divides its individuals between the two actions can reproduce in either environmental state. This is the probability-matching result of \citet{brennan2011origin}.

To compare how different evolutionary mechanisms adapt to a changing environment, we need a measure of how much growth a lineage loses when its behavior deviates from the optimum. The following proposition defines this loss and establishes its connection to information theory.

\begin{proposition}[Growth regret is relative entropy]\label{prop:kl}
For $p,f\in(0,1)$, the growth regret of behavior $f$ is
\begin{equation}
\Rcal_p(f)\equiv I_p(p)-I_p(f)
=p\log\frac{p}{f}+(1-p)\log\frac{1-p}{1-f}
=\KL\!\left(\mathrm{Bern}(p)\,\|\,\mathrm{Bern}(f)\right)\ge0,
\label{eq:kl}
\end{equation}
where $\mathrm{Bern}(\theta)$ denotes the Bernoulli distribution with success probability $\theta$, and $\KL$ denotes the KL divergence. Equality holds if and only if $f=p$.
\end{proposition}

The proposition shows that growth regret can be expressed exactly as the KL divergence between the environmental distribution and the inherited behavioral distribution. This result is closely related to the Kelly criterion \citep{kelly1956}, which characterizes the cost of betting according to incorrect probabilities. For a lineage with the behavior $f_t$ in generation $t$, $\Rcal_p(f_t)$ is the expected log growth that it loses from generation $t$ to generation $t+1$ relative to a lineage with the optimal behavior $p$. Growth regret thus provides a natural measure of maladaptation, allowing us to compare the three evolutionary mechanisms in terms of how quickly they recover the growth lost after an environmental change.

To understand the magnitude of growth regret and how it varies with behavior $f$, consider its Taylor expansion around the optimal behavior $f=p$:
\begin{equation}
\Rcal_p(f)
=\frac{(f-p)^2}{2p(1-p)}+O\big((f-p)^3\big),
\qquad f\to p.
\label{eq:localKL}
\end{equation}
The expansion shows that, for small deviations, growth regret increases approximately with the square of the distance from the optimum. Doubling the deviation therefore approximately quadruples the loss in growth.

\subsection{Growth of a population}\label{ssec:population}

So far, we have considered a single lineage whose individuals share the same behavior within each generation. We now extend the analysis to a population consisting of several lineages, each with its own behavior. For simplicity, we focus on two lineages, although the results extend to more. In generation $t$, let $f_{1,t}$ and $f_{2,t}$ denote their behaviors, $q_t$ and $1-q_t$ their population shares, and $N_t$ the size of the entire population. The two lineages therefore contain $q_tN_t$ and $(1-q_t)N_t$ individuals, respectively.

To determine the growth of the entire population, we first consider the offspring produced by each lineage. If $X_{t+1}=\alpha$, only individuals choosing action $a$ reproduce. The first lineage therefore grows by the factor $mf_{1,t}$, contributing $mf_{1,t}q_tN_t$ individuals to the next generation. Similarly, the second lineage contributes $mf_{2,t}(1-q_t)N_t$ individuals. Adding the two contributions, the population has $mf_{1,t}q_tN_t+mf_{2,t}(1-q_t)N_t=m\fbar_tN_t$ individuals in generation $t+1$, where
\begin{equation}
\fbar_t\equiv q_tf_{1,t}+(1-q_t)f_{2,t}
\label{eq:fbar}
\end{equation}
is the average behavior of the population. If $X_{t+1}=\beta$, only individuals choosing action $b$ reproduce. The two lineages then contribute $m(1-f_{1,t})q_tN_t$ and $m(1-f_{2,t})(1-q_t)N_t$ individuals, respectively, giving a total population of $m(1-\fbar_t)N_t$ in generation $t+1$.

We can also use these results to determine how the relative shares of the two lineages evolve through natural selection. The share of the first lineage in generation $t+1$ is its number of offspring divided by the total population size:
\begin{equation}
q_{t+1}=
\begin{cases}
\displaystyle\frac{mf_{1,t}q_tN_t}{m\fbar_tN_t}
=\frac{q_tf_{1,t}}{\fbar_t}, & X_{t+1}=\alpha,\\[8pt]
\displaystyle\frac{m(1-f_{1,t})q_tN_t}{m(1-\fbar_t)N_t}
=\frac{q_t(1-f_{1,t})}{1-\fbar_t}, & X_{t+1}=\beta.
\end{cases}
\label{eq:shares}
\end{equation}
Equation~\eqref{eq:shares} describes how natural selection changes the relative shares of the two lineages from one generation to the next.

We now consider the growth of the population as a whole. When $X_{t+1}=\alpha$, its size increases from $N_t$ to $m\fbar_tN_t$, whereas when $X_{t+1}=\beta$, it increases to $m(1-\fbar_t)N_t$. Taking the expected logarithm of the population growth ratio $N_{t+1}/N_t$ gives
\begin{equation}
I_p(q_t,f_{1,t},f_{2,t})
\equiv\E\Big[\log\frac{N_{t+1}}{N_t}\,\Big|\,q_t,f_{1,t},f_{2,t}\Big]=p\log(m\fbar_t)+(1-p)\log\big(m(1-\fbar_t)\big)=I_p(\fbar_t),
\label{eq:growthTwo}
\end{equation}
where the last equality follows from Equation~\eqref{eq:growth} with $\fbar_t$ in place of $f_t$. Equation~\eqref{eq:growthTwo} shows that the population's expected log growth depends on the behaviors of the two lineages only through their average $\fbar_t$. This is because reproductive success depends on the chosen action and the environmental state, rather than on which lineage an individual belongs to.

We can therefore measure the growth regret of the entire population in the same way as for a single lineage, by comparing its expected log growth with the optimal growth rate $I_p(p)$. Combining Equation~\eqref{eq:growthTwo} with \Cref{prop:kl} gives the following result.

\begin{proposition}[Growth regret of a population with two behaviors]\label{prop:average}
For $p,f_{1,t},f_{2,t}\in(0,1)$ and $q_t\in[0,1]$,
\[
I_p(p)-I_p(q_t,f_{1,t},f_{2,t})=\Rcal_p(\fbar_t).
\]
\end{proposition}

The population's growth regret is therefore determined by its average behavior $\fbar_t$.

\subsection{Mechanisms and measures of adaptation}\label{ssec:shift}

The preceding analysis shows that a population's growth regret depends on its average behavior, which can change through two channels: natural selection changes the relative shares of lineages, while learning or search can change their inherited behaviors. We consider three evolutionary mechanisms that differ in which of these channels are available. Under Darwinian evolution, inherited behaviors remain fixed, and adaptation occurs through natural selection alone. Lamarckian evolution additionally allows individuals to learn and pass their acquired improvements to their descendants. Agent-directed evolution further allows individuals to search among variants of their behavior. Natural selection operates under all three mechanisms.

To compare these mechanisms, we consider a common environmental change. Suppose that at time $0$, the probability of state $\alpha$ changes permanently from $p_{\mathrm{old}}$ to $p_{\mathrm{new}}\ne p_{\mathrm{old}}$. By \Cref{prop:pm}, the growth-optimal behavior changes from $p_{\mathrm{old}}$ to $p_{\mathrm{new}}$. We study how quickly each mechanism allows the population to adapt to the new environment.

We assume that all three mechanisms start from the same population. At time $0$, a fraction $q_0\in(0,1)$ of individuals already follows the new optimal behavior $p_{\mathrm{new}}$, while the remaining fraction $1-q_0$ follows the previously optimal behavior $p_{\mathrm{old}}$. We refer to their descendants as the \emph{adapted lineage} and the \emph{maladapted lineage}, respectively. These names describe their initial behaviors, since the maladapted lineage may become better adapted over time. The adapted lineage retains behavior $p_{\mathrm{new}}$ throughout, while the behavior of the maladapted lineage evolves according to the mechanism under consideration.

We denote the three mechanisms by $\mD$, $\mL$, and $\mA$, respectively. For each mechanism $\mM\in\{\mD,\mL,\mA\}$, let $q_t^{\mM}$ denote the population share of the adapted lineage and $f_t^{\mM}$ the behavior of the maladapted lineage in generation $t$. By Equation~\eqref{eq:fbar}, the population's average behavior is
\[
\fbar_t^{\mM}
=q_t^{\mM}p_{\mathrm{new}}
+(1-q_t^{\mM})f_t^{\mM}.
\]
Under all three mechanisms, natural selection changes the population share $q_t^{\mM}$ according to Equation~\eqref{eq:shares}, with $f_{1,t}=p_{\mathrm{new}}$ and $f_{2,t}=f_t^{\mM}$. The mechanisms differ in how the behavior $f_t^{\mM}$ evolves across generations. By \Cref{prop:average}, the population's growth regret in generation $t$ is $\Rcal_{p_{\mathrm{new}}}(\fbar_t^{\mM})$, which we use to measure maladaptation throughout. Since all three mechanisms share the initial conditions $q_0^{\mM}=q_0$ and $f_0^{\mM}=p_{\mathrm{old}}$, they also have the same initial growth regret. We can therefore compare how quickly the three mechanisms reduce growth regret from a common starting point.

There are two natural ways to measure the speed of adaptation. We can ask how many generations are needed to reduce regret by a specified fraction, or how rapidly regret declines over the long run. These lead to two complementary measures, adaptation time and adaptation rate, which we define for all three evolutionary mechanisms.

\begin{definition}[Adaptation time]\label{def:time}
For $\varepsilon\in(0,1)$ and $q_0\in[0,1)$, the adaptation time of Mechanism $\mM$ is
\[
\tau^{\mM}(\varepsilon,q_0)\equiv
\min\left\{t\ge0:\ \Rcal_{p_{\mathrm{new}}}(\fbar^{\mM}_t)
\le\varepsilon\,\Rcal_{p_{\mathrm{new}}}(\fbar^{\mM}_0)\right\},
\]
with the convention that $\min\varnothing=\infty$.
\end{definition}

\begin{definition}[Adaptation rate]\label{def:rate}
The adaptation rate of Mechanism $\mM$ is
\[
\kappa^{\mM}\equiv-\lim_{t\to\infty}\frac1t
\log\Rcal_{p_{\mathrm{new}}}(\fbar^{\mM}_t),
\]
whenever the limit exists with probability one.
\end{definition}

The adaptation time is the first generation in which growth regret falls to at most a fraction $\varepsilon$ of its initial value. It depends on the mechanism $\mM$, the threshold $\varepsilon$, and the initial share $q_0$ of the adapted lineage. The adaptation rate measures the long-run exponential decay of growth regret, with a larger rate indicating faster convergence.

\section{Darwinian Evolution}\label{sec:darwinian}

We begin with Darwinian evolution, in which the two lineages retain their respective behaviors $p_{\mathrm{new}}$ and $p_{\mathrm{old}}$ across generations. Since their behaviors remain fixed, the population can adapt only through changes in their relative shares under natural selection. This mechanism provides a benchmark for the other two. \Cref{ssec:darwinshift} introduces the Darwinian mechanism, and \Cref{ssec:darwinrate} analyzes its adaptation time and rate.

\subsection{The model}\label{ssec:darwinshift}

The defining feature of Darwinian evolution is that adaptation occurs through natural selection rather than changes in inherited behavior. The two lineages retain their respective behaviors $p_{\mathrm{new}}$ and $p_{\mathrm{old}}$, while differences in reproductive success change their relative population shares. Following Equation~\eqref{eq:shares}, Mechanism~\ref{mech:D} describes this process from one generation to the next.

\begin{mechanism}{D}{Darwinian evolution}{mech:D}
\item[Start.] Generation $0$ has $q^{\mD}_0=q_0$ and $f^{\mD}_0=p_{\mathrm{old}}$, and each generation $t\ge1$ is formed from generation $t-1$ in the steps below; throughout, $\fbar^{\mD}_t\equiv q^{\mD}_tp_{\mathrm{new}}+(1-q^{\mD}_t)f^{\mD}_t$.
\item[Inheritance.] $f^{\mD}_t=f^{\mD}_{t-1}=p_{\mathrm{old}}$.
\item[Selection.] Once the state $X_t$ is realized,
\[
q^{\mD}_t=
\begin{cases}
q^{\mD}_{t-1}\,p_{\mathrm{new}}/\fbar^{\mD}_{t-1}, & X_t=\alpha,\\[2pt]
q^{\mD}_{t-1}\,(1-p_{\mathrm{new}})/(1-\fbar^{\mD}_{t-1}), & X_t=\beta.
\end{cases}
\]
\end{mechanism}

By \Cref{prop:average}, the whole population grows in each generation as a single type-$\fbar^{\mD}_t$ lineage would, and its growth regret in generation $t$ is $\Rcal_{p_{\mathrm{new}}}(\fbar^{\mD}_t)$. Although the behaviors of the two lineages remain fixed, natural selection changes their relative shares and hence the population's average behavior.

\subsection{Adaptation time and rate}\label{ssec:darwinrate}

The following theorem characterizes the adaptation time $\tau^{\mD}(\varepsilon,q_0)$ and rate $\kD$ of Mechanism~\ref{mech:D}, as defined in \Cref{def:time,def:rate}.

\begin{theorem}[Darwinian adaptation rate and time]\label{thm:darwin}
Under Mechanism~\ref{mech:D}:
\begin{enumerate}[label=(\roman*)]
\item for every $q_0\in(0,1)$, $\kD=2\Delta$ with probability one;
\item
\begin{equation}
\begin{aligned}
&\lim_{\varepsilon\to0}\frac{\E\,\tau^{\mD}(\varepsilon,q_0)}{\log(1/\varepsilon)}=\frac1{\kD}&&\text{for every }q_0\in(0,1),\\
&\lim_{q_0\to0}\frac{\E\,\tau^{\mD}(\varepsilon,q_0)}{\log(1/q_0)}=\frac1\Delta&&\text{for every }\varepsilon\in(0,1).
\end{aligned}
\label{eq:darwinTime}
\end{equation}
\end{enumerate}
Here,
\begin{equation}
\Delta\equiv
p_{\mathrm{new}}\log\frac{p_{\mathrm{new}}}{p_{\mathrm{old}}}
+(1-p_{\mathrm{new}})\log\frac{1-p_{\mathrm{new}}}{1-p_{\mathrm{old}}}
=\Rcal_{p_{\mathrm{new}}}(p_{\mathrm{old}})>0.
\label{eq:Delta}
\end{equation}
\end{theorem}

Part (i) of \Cref{thm:darwin} shows that the long-run adaptation rate under Darwinian evolution is $\kD=2\Delta$. The quantity $\Delta$ is the adapted lineage's advantage in expected log growth per generation. The larger the growth advantage $\Delta$ of the adapted lineage, the faster it spreads through the population and the faster growth regret declines. The factor of two arises because growth regret is approximately quadratic near the optimum (Equation~\eqref{eq:localKL}). Importantly, the adaptation rate does not depend on the initial share $q_0$ of the adapted lineage: regardless of how rare it is initially, the population eventually adapts at the same long-run rate.

Part (ii) of \Cref{thm:darwin} explains how the expected adaptation time depends on the threshold $\varepsilon$ and the initial share $q_0$. First, for any fixed $q_0>0$, the expected adaptation time grows like $\log(1/\varepsilon)/\kD$ as $\varepsilon$ approaches zero. Second, for any fixed threshold $\varepsilon$, the expected adaptation time grows like $\log(1/q_0)/\Delta$ as $q_0$ approaches zero. The intuition is that when the adapted lineage is initially rare, it must first grow to a sufficiently large share before it can substantially improve the population's average behavior. We call this additional delay the \emph{seed penalty}. It can become arbitrarily large even though the long-run adaptation rate remains unchanged. At $q_0=0$, adaptation is impossible: without any individuals carrying the adapted behavior, natural selection has no way to introduce it.\footnote{Mutation could introduce new behaviors and thereby make adaptation possible, but we do not consider mutation in this paper.}

Darwinian adaptation operates entirely through natural selection among fixed inherited behaviors. We next consider Lamarckian evolution, in which learning can also change the behavior passed on to future generations.

\section{Lamarckian Evolution}\label{sec:lamarckian}

We now consider Lamarckian evolution, which adds the inheritance of acquired improvements to the Darwinian framework: natural selection still determines the relative shares of the two lineages, but the maladapted lineage can also learn from the new environment and pass its improved behavior to its descendants. \Cref{ssec:lamrule} introduces the Lamarckian mechanism and examines how learning interacts with selection, and \Cref{ssec:lamrate} analyzes its adaptation time and rate.

\subsection{The model}\label{ssec:lamrule}

Under Lamarckian evolution, individuals can acquire better behavior during their lifetimes and pass it on to their descendants, so the initially maladapted lineage can improve its inherited behavior over successive generations instead of relying entirely on natural selection to increase the share of the already adapted lineage.

How much can learning improve inherited behavior in each generation? If individuals could fully adopt the new optimal behavior $p_{\mathrm{new}}$ and transmit it to their descendants, they could reach the optimum in a single generation. However, individuals may be able to change their behavior only partly within a lifetime, and acquired improvements may not be fully transmitted.
We capture these limitations by assuming that, in each generation, learning and transmission move the maladapted lineage's behavior a fixed fraction of the way toward the new optimum:
\begin{equation}
f^{\mL}_{t+1}=f^{\mL}_t+\lambda\,(p_{\mathrm{new}}-f^{\mL}_t),
\qquad \lambda\in(0,1),
\label{eq:lamUpdate}
\end{equation}
with $f^{\mL}_0=p_{\mathrm{old}}$.

The parameter $\lambda$ measures the effectiveness of inherited learning: it combines how much individuals can change their behavior within a generation with how faithfully these changes are passed on. A larger $\lambda$ closes more of the remaining gap in each generation, and in the limiting case $\lambda=1$, the lineage would reach the new optimum in a single generation.

Equation~\eqref{eq:lamUpdate} specifies the direction and speed of learning rather than how individuals learn. Behavior moves toward the optimum of the new environment, much as training moves an AI system toward the behavior that its objective rewards. This adjustment rule is closely related to partial-adjustment and adaptive-expectations models in economics \citep{cagan1956monetary,nerlove1958adaptive}.

Combining learning with the selection rule in Equation~\eqref{eq:shares} gives Mechanism~\ref{mech:L}.

\begin{mechanism}{L}{Lamarckian evolution}{mech:L}
\item[Start.] Generation $0$ has $q^{\mL}_0=q_0$ and $f^{\mL}_0=p_{\mathrm{old}}$, and each generation $t\ge1$ is formed from generation $t-1$ in the steps below; throughout, $\fbar^{\mL}_t\equiv q^{\mL}_tp_{\mathrm{new}}+(1-q^{\mL}_t)f^{\mL}_t$.
\item[Learning.] $f^{\mL}_t=f^{\mL}_{t-1}+\lambda\,(p_{\mathrm{new}}-f^{\mL}_{t-1})$.
\item[Selection.] Once the state $X_t$ is realized,
\[
q^{\mL}_t=
\begin{cases}
q^{\mL}_{t-1}\,p_{\mathrm{new}}/\fbar^{\mL}_{t-1}, & X_t=\alpha,\\[2pt]
q^{\mL}_{t-1}\,(1-p_{\mathrm{new}})/(1-\fbar^{\mL}_{t-1}), & X_t=\beta.
\end{cases}
\]
\end{mechanism}

Mechanism~\ref{mech:L} shows the two channels of adaptation directly: $\fbar^{\mL}_t-p_{\mathrm{new}}=(1-q^{\mL}_t)(f^{\mL}_t-p_{\mathrm{new}})$. Natural selection changes the population share $1-q^{\mL}_t$ of the maladapted lineage, while learning eliminates a fraction $\lambda$ of its remaining behavioral gap $f^{\mL}_t-p_{\mathrm{new}}$ in each generation. The population can thus approach the new optimum even without the adapted lineage taking over.

\subsection{Adaptation time and rate}\label{ssec:lamrate}

The following theorem characterizes the adaptation time $\tau^{\mL}(\varepsilon,q_0)$ and rate $\kL$ of Mechanism~\ref{mech:L}, as defined in \Cref{def:time,def:rate}.

\begin{theorem}[Lamarckian adaptation rate and time]\label{thm:lam}
Under Mechanism~\ref{mech:L}:
\begin{enumerate}[label=(\roman*)]
\item for every $q_0\in[0,1)$, $\kL=-2\log(1-\lambda)$ along every history;
\item
\[
\begin{aligned}
&\lim_{\varepsilon\to0}\frac{\E\,\tau^{\mL}(\varepsilon,q_0)}{\log(1/\varepsilon)}=\frac1{\kL}&&\text{for every }q_0\in[0,1),\\
&\sup_{q_0\in[0,1)}\E\,\tau^{\mL}(\varepsilon,q_0)<\infty&&\text{for every }\varepsilon\in(0,1).
\end{aligned}
\]
\end{enumerate}
\end{theorem}

Part (i) of \Cref{thm:lam} shows that the long-run adaptation rate under Lamarckian evolution is $\kL=-2\log(1-\lambda)$: the larger the learning parameter $\lambda$, the faster the maladapted lineage moves toward the new optimum. Unlike under Darwinian evolution, the rate is determined by the speed of learning rather than by the growth advantage $\Delta$ of the adapted lineage. Under Darwinian evolution, the two lineages retain their original behaviors, so the adapted lineage keeps a constant growth advantage. Under Lamarckian evolution, learning gradually eliminates this advantage by improving the maladapted lineage, and natural selection weakens as the two lineages become more similar. Although selection continues to operate, long-run adaptation is therefore driven by learning rather than by changes in population shares.

Part (ii) of \Cref{thm:lam} characterizes the expected adaptation time. For any fixed $q_0\in[0,1)$, it grows like $\log(1/\varepsilon)/\kL$ as $\varepsilon$ approaches zero. More importantly, it remains uniformly bounded across all initial shares $q_0\in[0,1)$. This is a fundamental difference from Darwinian evolution, where an initially rare adapted lineage can lead to an arbitrarily large seed penalty (\Cref{thm:darwin}). Under Lamarckian evolution, the maladapted lineage can improve its own behavior rather than wait to be replaced, so even when $q_0=0$, with no individual initially possessing the optimal behavior, learning can still bring the population toward the new optimum.

Under Mechanism~\ref{mech:L}, however, the learning process is fixed: $\lambda$ determines the fraction of the remaining gap eliminated in each generation, and the system does not decide how to search for improvements. We next consider agent-directed evolution, in which the system can generate and evaluate multiple variants of itself and choose how many to explore, so that the process of adaptation itself becomes a decision.

\section{Agent-Directed Evolution}\label{sec:agentic}

We now turn to agent-directed evolution, which adds the ability to search for improvements to the Lamarckian framework: in addition to learning, the maladapted lineage can generate multiple variants of its behavior, evaluate them, and retain the best one for further learning. More importantly, the agent can decide how many variants to explore, which lets it influence the process by which it adapts. \Cref{ssec:heterogeneity} introduces the agent-directed mechanism, \Cref{ssec:commonstate} analyzes its adaptation time and rate and the gain from search, and \Cref{ssec:choiceN} examines how many variants the agent chooses when it maximizes the speed of adaptation and when it minimizes its cost.

\subsection{The model}\label{ssec:heterogeneity}

In Mechanism~\ref{mech:L}, the maladapted lineage improves its inherited behavior through a fixed learning rule. An intelligent agent may also be able to experiment with different ways of improving its behavior before passing it on: it can generate several candidate modifications of its current behavior, evaluate them, and retain the best one for further learning. We call these candidates \emph{variants} and let $n\ge1$ denote the number of variants the agent chooses to generate in each generation.

How much improvement can these variants achieve? The outcome of any attempt is uncertain, with some variants producing substantial improvements and others very little, and the difficulty of the problem may limit how much progress even the best attempt can make. We capture both features by letting each variant move toward the new optimum by a random fraction of the remaining behavioral gap: in generation $t$, the agent generates $n$ variants with behaviors
\begin{equation}
x_{t,j}=f^{\mA}_{t-1}+\rho\,U_{t,j}\,(p_{\mathrm{new}}-f^{\mA}_{t-1}),
\qquad j=1,\ldots,n,
\label{eq:variant}
\end{equation}
where the random variables $U_{t,j}$ are independent, uniformly distributed on $(0,1)$, and independent of the environment.

The parameter $\rho\in(0,1]$ represents the difficulty of the problem: it is the maximum fraction of the remaining gap that a single variant can eliminate. A smaller $\rho$ corresponds to a harder problem, in which even a successful attempt makes only limited progress, and when $\rho=1$, a single variant can move arbitrarily close to the optimum. The agent can choose how many variants to generate, but it controls neither their random outcomes nor $\rho$. As with the Lamarckian learning rule, Equation~\eqref{eq:variant} is a reduced-form assumption: the agent need not know $p_{\mathrm{new}}$, but its attempts are assumed to move in the correct direction, with uncertain magnitudes.

We assume that an external evaluator compares the variants and identifies the one whose behavior is closest to $p_{\mathrm{new}}$; the agent itself need not know $p_{\mathrm{new}}$ and only receives the evaluator's assessment. The selected variant then undergoes the same learning as in Mechanism~\ref{mech:L}, moving a fraction $\lambda$ of the remaining gap toward the optimum, and natural selection operates between the two lineages as before. Combining these steps gives Mechanism~\ref{mech:A}.

\begin{mechanism}{A}{agent-directed evolution}{mech:A}
\item[Start.] Generation $0$ has $q^{\mA}_0=q_0$ and $f^{\mA}_0=p_{\mathrm{old}}$, and each generation $t\ge1$ is formed from generation $t-1$ in the steps below; throughout, $\fbar^{\mA}_t\equiv q^{\mA}_tp_{\mathrm{new}}+(1-q^{\mA}_t)f^{\mA}_t$.
\item[Search.] The agent creates $n$ variants with behaviors $x_{t,j}=f^{\mA}_{t-1}+\rho\,U_{t,j}(p_{\mathrm{new}}-f^{\mA}_{t-1})$ for $j=1,\ldots,n$, where the draws $U_{t,j}$ are independent and uniform on $(0,1)$.
\item[Evaluation.] The evaluator reports $J_t\in\arg\min_{1\le j\le n}|x_{t,j}-p_{\mathrm{new}}|$, and variant $J_t$ is kept.
\item[Learning.] $f^{\mA}_t=x_{t,J_t}+\lambda\,(p_{\mathrm{new}}-x_{t,J_t})$.
\item[Selection.] Once the state $X_t$ is realized,
\[
q^{\mA}_t=
\begin{cases}
q^{\mA}_{t-1}\,p_{\mathrm{new}}/\fbar^{\mA}_{t-1}, & X_t=\alpha,\\[2pt]
q^{\mA}_{t-1}\,(1-p_{\mathrm{new}})/(1-\fbar^{\mA}_{t-1}), & X_t=\beta.
\end{cases}
\]
\end{mechanism}

Since every variant moves toward the new optimum, the evaluator selects the variant with the largest random draw, $U^*_t\equiv\max\{U_{t,1},\ldots,U_{t,n}\}$. Search eliminates a fraction $\rho U^*_t$ of the remaining behavioral gap, and learning then eliminates a fraction $\lambda$ of what remains. Combining these two steps gives
\begin{equation}
f^{\mA}_t-p_{\mathrm{new}}
=(1-\lambda)(1-\rho U^*_t)(f^{\mA}_{t-1}-p_{\mathrm{new}}).
\label{eq:agentStep}
\end{equation}
If the agent generates no variants ($n=0$), we set $U^*_t=0$, and Equation~\eqref{eq:agentStep} reduces to the Lamarckian learning rule in Equation~\eqref{eq:lamUpdate}: Mechanism~\ref{mech:L} is the special case of Mechanism~\ref{mech:A} without search.

\subsection{Adaptation time and rate}\label{ssec:commonstate}

The following theorem gives the adaptation rate and time of Mechanism~\ref{mech:A}, as defined in \Cref{def:time,def:rate}.

\begin{theorem}[Agent-directed adaptation rate and time]\label{thm:agent}
Under Mechanism~\ref{mech:A}:
\begin{enumerate}[label=(\roman*)]
\item for every $q_0\in[0,1)$, with probability one,
\[
\kA=\kL+2G_n(\rho),
\]
where
\begin{equation}
G_n(\rho)\equiv\E\big[-\log(1-\rho\,U^*_t)\big]\ge0\quad\text{for }n\ge1,\qquad G_0(\rho)=0;
\label{eq:GN}
\end{equation}
\item
\begin{equation}
\begin{aligned}
&\lim_{\varepsilon\to0}\frac{\E\,\tau^{\mA}(\varepsilon,q_0)}{\log(1/\varepsilon)}=\frac1{\kA}&&\text{for every }q_0\in[0,1),\\
&\sup_{q_0\in[0,1)}\E\,\tau^{\mA}(\varepsilon,q_0)<\infty&&\text{for every }\varepsilon\in(0,1).
\end{aligned}
\label{eq:agentTime}
\end{equation}
\end{enumerate}
\end{theorem}

Part (i) of \Cref{thm:agent} shows that the long-run adaptation rate under agent-directed evolution is $\kA=\kL+2G_n(\rho)$: unlike the Lamarckian rate, which depends only on the learning parameter $\lambda$, it benefits from both learning and search. The term $G_n(\rho)$ measures the expected improvement from the best of $n$ variants. As in Lamarckian evolution, natural selection continues to operate, but its influence gradually diminishes as the two lineages become better adapted, so the long-run rate is determined by learning and search.

Part (ii) of \Cref{thm:agent} characterizes the expected adaptation time. For any fixed $q_0\in[0,1)$, it grows like $\log(1/\varepsilon)/\kA$ as $\varepsilon$ approaches zero, and it remains uniformly bounded across all initial shares $q_0\in[0,1)$. Like Lamarckian evolution, agent-directed evolution therefore has no seed penalty, but search provides an additional source of improvement that lets the agent adapt faster than through learning alone.

The following proposition characterizes how the gain from search depends on the number of variants $n$ and the difficulty parameter $\rho$.

\begin{proposition}[What search adds]\label{prop:gain}
For the function $G_n(\rho)$ in Equation~\eqref{eq:GN}:
\begin{enumerate}[label=(\roman*)]
\item $G_n(\rho)$ is strictly increasing in $n$ for every $\rho\in(0,1]$, and strictly increasing in $\rho$ for every $n\ge1$;
\item $G_n(1)=1+\frac12+\cdots+\frac1n$, and $\lim_{n\to\infty}G_n(1)/\log n=1$;
\item for $\rho\in(0,1)$, $G_n(\rho)<-\log(1-\rho)$ for every $n$, and $\lim_{n\to\infty}G_n(\rho)=-\log(1-\rho)$.
\end{enumerate}
\end{proposition}

Part (i) of \Cref{prop:gain} shows that the gain from search increases with both the number of variants $n$ and the difficulty parameter $\rho$: more variants raise the chance of finding one that makes substantial progress toward the optimum, and a larger $\rho$ allows each variant to achieve greater improvements.

Part (ii) of \Cref{prop:gain} considers the case $\rho=1$, where a single variant can move arbitrarily close to the optimum. Then $G_n(1)=1+\frac12+\cdots+\frac1n$ grows approximately like $\log n$, so the additional adaptation rate from search grows approximately as $2\log n$: each additional variant contributes less than the previous one, but the total gain increases without bound.

Part (iii) of \Cref{prop:gain} considers the case $\rho<1$, where each variant can eliminate at most a fraction $\rho$ of the remaining behavioral gap. As the number of variants increases, the best variant approaches this maximum and the adaptation rate approaches $\kL-2\log(1-\rho)$, so, unlike when $\rho=1$, the gain from search has a finite upper bound that depends on the difficulty of the problem. \Cref{fig:gain} in Appendix~\ref{app:figures} illustrates \Cref{prop:gain}.

\subsection{Choosing the number of variants}\label{ssec:choiceN}

So far, we have treated the number of variants $n$ as given. An agent that can choose how many variants to generate must decide how much search to undertake, and this decision depends on what it seeks to achieve. We consider two objectives: maximizing the speed of adaptation and minimizing its cost.

\paragraph{Speed-maximizing agent.}
Consider first an agent whose primary objective is to adapt as quickly as possible, without accounting for the cost of generating variants. It chooses $n$ to maximize its adaptation rate $\kA=\kL+2G_n(\rho)$, which increases with $n$ for every $\rho\in(0,1]$ by \Cref{prop:gain}(i). Regardless of the difficulty of the problem, more variants therefore always lead to faster adaptation: the gain from search increases at a diminishing rate and may approach a finite upper bound when $\rho<1$, but every additional variant still contributes. In the absence of resource constraints, a speed-maximizing agent has no finite optimal $n$ and would generate as many variants as its resources permit.

OpenAI's use of roughly 10,000 agents to tackle the Navier--Stokes problem, as discussed in the Introduction, illustrates this approach \citep{openai2026navier}. The scale of this effort suggests a willingness to commit substantial computational resources to exploring more possibilities in parallel, prioritizing rapid progress over minimizing the cost of search. When speed is the primary objective, generating more variants can therefore be worthwhile even at considerable expense. We next consider how the agent's choice changes when minimizing the total cost of adaptation becomes the objective.

\paragraph{Cost-minimizing agent.} A different objective arises for AI agents deployed to perform large numbers of tasks, such as automated bug fixing or code optimization, which may seek to reach a given performance target at the lowest possible computational cost. More variants reduce the number of generations needed to reach the target but increase the cost of each generation. To capture this tradeoff, suppose that learning costs one unit per generation and each variant an additional $c>0$, so that a generation with $n$ variants costs $1+cn$.

The total cost of adaptation is the cost per generation times the number of generations needed to adapt. By \Cref{thm:agent}(ii), for sufficiently stringent adaptation targets, the expected number of generations is approximately proportional to $1/\kA$. We therefore define the \emph{long-run cost} of adaptation, approximately proportional to the expected total cost, as
\begin{equation}
C(n)\equiv\frac{1+cn}{\kA}=\frac{1+cn}{\kL+2G_n(\rho)}.
\label{eq:longRunCost}
\end{equation}
Assuming that the agent knows its learning parameter $\lambda$, the difficulty parameter $\rho$, and the cost $c$, it chooses the number of variants to minimize this long-run cost:
\begin{equation}
n^*(\rho,\lambda,c)\in\arg\min_{n\in\{0,1,2,\ldots\}}C(n).
\label{eq:Nchoice}
\end{equation}

The following proposition characterizes the optimal number of variants and how it depends on these parameters.

\begin{proposition}[The optimal number of variants]\label{prop:Nstar}
For the long-run cost $C(n)$ in Equation~\eqref{eq:longRunCost} and the choice of $n$ in Equation~\eqref{eq:Nchoice}:
\begin{enumerate}[label=(\roman*)]
\item $C(n)$ first decreases and then increases in $n$;
\item $n^*(\rho,\lambda,c)$ is nondecreasing in $\rho$ and nonincreasing in $\lambda$ and in $c$.
\end{enumerate}
\end{proposition}

Part (i) of \Cref{prop:Nstar} shows how the agent balances the benefits and costs of additional variants. Each new variant increases the adaptation rate by less as the number of variants grows, while its cost remains constant. When search is worthwhile, increasing $n$ therefore first reduces the long-run cost, but eventually the additional cost outweighs the benefit of faster adaptation, so the agent chooses the number of variants that minimizes $C(n)$ rather than generating as many as possible.

Part (ii) of \Cref{prop:Nstar} shows how this choice depends on the agent's learning ability, the difficulty of the problem, and the cost of search. A larger $\lambda$ lets the agent adapt more quickly through learning alone, reducing the need for search, and a higher $c$ makes variants less attractive, whereas a larger $\rho$ allows each variant to achieve greater improvements and makes search more valuable. The agent therefore chooses more variants when it learns more slowly, when search is less expensive, and when the problem is easier. \Cref{fig:evolvability} in Appendix~\ref{app:figures} illustrates \Cref{prop:Nstar}.

We conclude this section by clarifying the distinction between maximizing adaptation speed and minimizing adaptation cost. The speed-maximizing agent maximizes the adaptation rate per generation. When variants can be generated and evaluated in parallel, as in OpenAI's example, additional variants need not lengthen a generation, so maximizing the rate per generation also maximizes the rate per unit of time. If variants are instead evaluated sequentially and the duration of each generation is proportional to $1+cn$, the adaptation rate per unit of time becomes $\kA/(1+cn)=1/C(n)$, and maximizing speed yields the same optimal choice $n^*$ as minimizing cost. The optimal search strategy thus depends not only on the agent's objective but also on whether additional variants can be evaluated in parallel.

\section{Comparing the Three Mechanisms}\label{sec:combined}

Having examined the three evolutionary mechanisms separately, we now compare their adaptation speeds. \Cref{ssec:hierarchy} compares their adaptation rates and times, and \Cref{sec:sims} illustrates the comparison numerically.

\subsection{Comparing adaptation speeds}\label{ssec:hierarchy}

The following theorem compares the adaptation speeds of the three evolutionary mechanisms. 

\begin{theorem}[Ranking of adaptation speeds]\label{thm:hierarchy}
Under Mechanisms~\ref{mech:D}, \ref{mech:L}, and~\ref{mech:A}, with $\rho\in(0,1]$, $n\ge1$, and $\lambda\in(0,1)$ satisfying
\begin{equation}
\lambda>1-e^{-\Delta},
\label{eq:speedCondition}
\end{equation}
\begin{enumerate}[label=(\roman*)]
\item for every $q_0\in(0,1)$, with probability one,
\[
\kD<\kL<\kA;
\]
\item for every $q_0\in(0,1)$, there exists $\bar\varepsilon>0$ such that
\begin{equation}
\E\,\tau^{\mA}(\varepsilon,q_0)<\E\,\tau^{\mL}(\varepsilon,q_0)<\E\,\tau^{\mD}(\varepsilon,q_0)
\qquad\text{for every }\varepsilon\in(0,\bar\varepsilon).
\label{eq:timeRanking}
\end{equation}
\end{enumerate}
\end{theorem}

Part (i) of \Cref{thm:hierarchy} shows that Lamarckian evolution adapts faster than Darwinian evolution when Equation~\eqref{eq:speedCondition} holds. As learning brings the two lineages closer in behavior, their growth rates converge and the influence of natural selection gradually diminishes, so long-run adaptation is driven by learning, which under this condition is faster than natural selection alone.

The second inequality in part (i) shows that agent-directed evolution adapts faster than Lamarckian evolution, because search adds $2G_n(\rho)$ to the Lamarckian adaptation rate (\Cref{thm:agent}).

Part (ii) of \Cref{thm:hierarchy} compares the expected times the three mechanisms need to reach the same adaptation target: under Equation~\eqref{eq:speedCondition}, for any initial share $q_0\in(0,1)$ and a sufficiently stringent target, agent-directed evolution requires the fewest generations on average, followed by Lamarckian and then Darwinian evolution.

\subsection{Numerical illustrations}\label{sec:sims}

We now illustrate the differences in adaptation speed among the three mechanisms numerically. Consider an environment in which the probability of state $\alpha$ changes permanently from $p_{\mathrm{old}}=0.40$ to $p_{\mathrm{new}}=0.60$. We set the initial share of the adapted lineage to $q_0=10^{-4}$ and the learning parameter to $\lambda=0.10$. Under agent-directed evolution, the agent generates one variant per generation ($n=1$), with the difficulty parameter $\rho=0.5$. These parameters give $\Delta=0.08109$, $\kD=0.16219$, $\kL=0.21072$, and $\kA=0.82443$. Since $\lambda=0.10>1-e^{-\Delta}=0.078$, the condition in Equation~\eqref{eq:speedCondition} holds, and the three mechanisms satisfy the adaptation rate ranking in \Cref{thm:hierarchy}.

\begin{figure}[htbp]
\centering
\begin{subfigure}[t]{.48\linewidth}
\includegraphics[width=\textwidth]{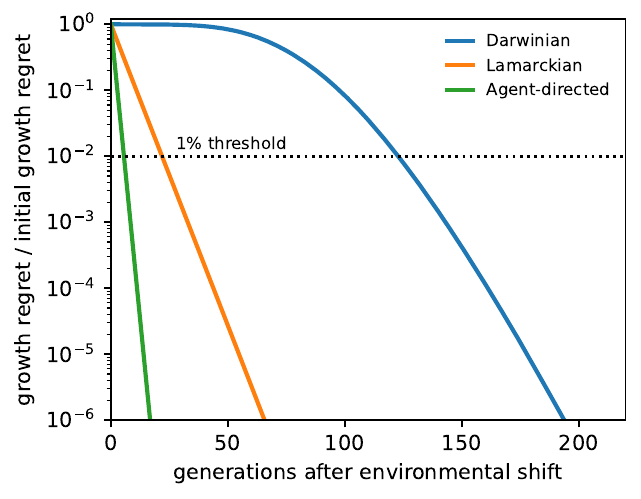}
\caption{Growth regret after the shift.\label{fig:speedPaths}}
\end{subfigure}\hfill
\begin{subfigure}[t]{.48\linewidth}
\includegraphics[width=\textwidth]{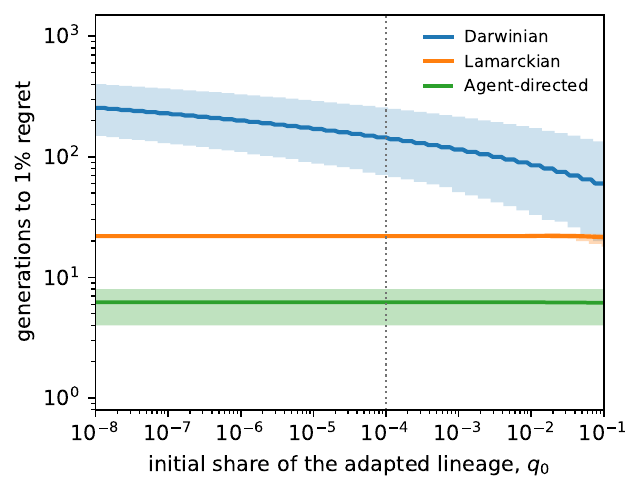}
\caption{Effect of the initial share $q_0$.\label{fig:speedSeed}}
\end{subfigure}

\caption{\textbf{Adaptation speed after an environmental shift.} Unless otherwise stated, $p_{\mathrm{old}}=0.40$, $p_{\mathrm{new}}=0.60$, $q_0=10^{-4}$, $\lambda=0.10$, and, for agent-directed evolution, $n=1$ and $\rho=0.5$. Panel (a) plots the geometric mean of each population's growth regret relative to its initial value, computed by averaging the logarithm of the regret ratio across the 200,000 simulated paths and then exponentiating. This preserves the long-run adaptation rates derived in \Cref{thm:darwin,thm:lam,thm:agent}. The dotted horizontal line marks one percent of initial regret. Panel (b) shows the expected number of generations needed to reach this threshold as $q_0$ varies. Shaded bands represent the 5th to 95th percentiles of adaptation times, and dotted vertical lines mark the baseline parameter values. All results are based on 200,000 simulated histories.}
\label{fig:speed}
\end{figure}

We conduct 200,000 Monte Carlo simulations, in which environmental states and the random improvements generated by the agent are drawn according to the model. \Cref{fig:speed} compares the adaptation of the three mechanisms and examines how their adaptation times depend on the initial share of the adapted lineage; \Cref{fig:speedMore} in Appendix~\ref{app:figures} shows how they depend on the learning parameter and the number of variants.

\Cref{fig:speedPaths} shows how growth regret declines under the three mechanisms. Under Darwinian evolution, regret remains close to its initial level for many generations because the adapted lineage is initially only a small fraction of the population. Under Lamarckian evolution, the maladapted lineage can improve its inherited behavior through learning, so the population adapts without waiting for natural selection to increase the share of the adapted lineage. Agent-directed evolution is faster still, as search improves the inherited behavior before learning takes place. Even with only one variant per generation, the additional improvement from search produces a substantial increase in adaptation speed.

\Cref{fig:speedSeed} examines the role of the initial share $q_0$. As the adapted lineage becomes rarer, Darwinian evolution requires increasingly many generations to reach the adaptation target, whereas the expected adaptation times under Lamarckian and agent-directed evolution remain relatively stable, even for very small $q_0$.

To compare the three mechanisms more directly, \Cref{tab:simtimes} reports their expected adaptation times for different regret thresholds. Under the baseline parameters, reducing growth regret to one percent of its initial value requires approximately 145 generations under Darwinian evolution, 22 under Lamarckian evolution, and 6.2 under agent-directed evolution with a single variant. When the threshold is reduced to 0.1 percent, the corresponding times are 160, 33, and 9.0 generations. These results illustrate the ranking in \Cref{thm:hierarchy}: adaptation times differ substantially even when the agent generates only a few variants.

\begin{table}[htbp]
\centering
\caption{Expected number of generations required to reduce growth regret to a given fraction of its initial value, based on 200,000 simulated histories. The parameters are $p_{\mathrm{old}}=0.40$, $p_{\mathrm{new}}=0.60$, $q_0=10^{-4}$, $\lambda=0.10$, and $\rho=0.5$.}
\label{tab:simtimes}
\small
\begin{tabular}{lrrrrrr}
\toprule
 & & & \multicolumn{4}{c}{Agent-directed} \\
\cmidrule(l){4-7}
Remaining regret & Darwinian & Lamarckian & $n=1$ & $n=2$ & $n=5$ & $n=10$ \\
\midrule
10\%  & 125 & 11 & 3.4 & 2.7 & 2.2 & 2.0 \\
5\%   & 130 & 15 & 4.3 & 3.4 & 2.9 & 2.7 \\
1\%   & 145 & 22 & 6.2 & 4.9 & 4.1 & 3.9 \\
0.1\% & 160 & 33 & 9.0 & 7.1 & 5.9 & 5.2 \\
\bottomrule
\end{tabular}
\end{table}

\section{Implications for AI Safety}\label{sec:safety}

We have shown that learning and search can substantially accelerate adaptation and that an agent can influence its own adaptation speed. This raises an important question for AI safety: does faster adaptation necessarily bring behavior closer to what the system's developers and users intend? The answer depends on how improvements are evaluated. In practice, AI systems are assessed using tests, benchmarks, reward models, and other criteria specified by humans, which may not fully capture the intended objectives, particularly when certain safety concerns are difficult to anticipate or measure. We now examine what happens when an agent adapts toward an incorrectly specified target and discuss the implications for evaluating and overseeing increasingly adaptive AI systems.

\subsection{Adaptation toward the wrong target}\label{ssec:pressure}

So far, we have assumed that search and learning move the agent's behavior toward the true optimum $p_{\mathrm{new}}$. If the evaluation criterion does not accurately reflect the true objective, however, the improvements the agent identifies may not be genuine improvements in the environment, and the agent may become increasingly successful according to its evaluator without necessarily achieving the behavior its developers intended.

To study this possibility, suppose that the agent's improvement is guided by a target $\tilde{p}\in(0,1)$ with $\tilde{p}\ne p_{\mathrm{new}}$; the difference between $\tilde{p}$ and $p_{\mathrm{new}}$ represents a systematic error in the evaluation criterion. The evaluator selects the variant closest to $\tilde{p}$ rather than $p_{\mathrm{new}}$, and search and learning also move the agent's behavior toward $\tilde{p}$. Natural selection, however, continues to operate through realized reproductive success in the true environment, where $p_{\mathrm{new}}$ remains optimal, so it continues to favor the adapted lineage with behavior $p_{\mathrm{new}}$. Mechanism~\ref{mech:F} describes this process.

\begin{mechanism}{$\mF$}{agent-directed evolution with a misspecified target}{mech:F}
\item[Start.] Generation $0$ has $q^{\mF}_0=q_0$ and $f^{\mF}_0=p_{\mathrm{old}}$, and each generation $t\ge1$ is formed from generation $t-1$ in the steps below; throughout, $\fbar^{\mF}_t\equiv q^{\mF}_tp_{\mathrm{new}}+(1-q^{\mF}_t)f^{\mF}_t$.
\item[Search.] The agent creates $n$ variants with behaviors $x_{t,j}=f^{\mF}_{t-1}+\rho\,U_{t,j}(\tilde{p}-f^{\mF}_{t-1})$ for $j=1,\ldots,n$, where the draws $U_{t,j}$ are independent and uniform on $(0,1)$.
\item[Evaluation.] The evaluator reports $J_t\in\arg\min_{1\le j\le n}|x_{t,j}-\tilde{p}|$, and variant $J_t$ is kept.
\item[Learning.] $f^{\mF}_t=x_{t,J_t}+\lambda\,(\tilde{p}-x_{t,J_t})$.
\item[Selection.] Once the state $X_t$ is realized,
\[
q^{\mF}_t=
\begin{cases}
q^{\mF}_{t-1}\,p_{\mathrm{new}}/\fbar^{\mF}_{t-1}, & X_t=\alpha,\\[2pt]
q^{\mF}_{t-1}\,(1-p_{\mathrm{new}})/(1-\fbar^{\mF}_{t-1}), & X_t=\beta.
\end{cases}
\]
\end{mechanism}

When $\tilde{p}=p_{\mathrm{new}}$, Mechanism~\ref{mech:F} reduces to Mechanism~\ref{mech:A}. When the two targets differ, progress according to the evaluator may no longer correspond to genuine adaptation. The following theorem characterizes the agent's adaptation toward the evaluator's target, its true growth regret, and the growth regret of the population as a whole.

\begin{theorem}[Adaptation toward a misspecified target]\label{thm:pressure}
Under Mechanism~\ref{mech:F}, with $\tilde{p}\in(0,1)\setminus\{p_{\mathrm{old}},p_{\mathrm{new}}\}$, $n\ge1$, and $q_0\in(0,1)$:
\begin{enumerate}[label=(\roman*)]
\item with probability one,
\[
-\lim_{t\to\infty}\frac1t\log\Rcal_{\tilde{p}}(f^{\mF}_t)
=\kL+2G_n(\rho);
\]
\item
\[
\lim_{t\to\infty}\Rcal_{p_{\mathrm{new}}}(f^{\mF}_t)
=\Rcal_{p_{\mathrm{new}}}(\tilde{p})>0;
\]
\item with probability one,
\[
-\lim_{t\to\infty}\frac1t\log\Rcal_{p_{\mathrm{new}}}(\fbar^{\mF}_t)
=2\Rcal_{p_{\mathrm{new}}}(\tilde{p}).
\]
\end{enumerate}
\end{theorem}

Part (i) of \Cref{thm:pressure} shows that the agent continues to adapt rapidly according to its evaluator, even when the evaluation target is incorrect: measured relative to $\tilde{p}$, its growth regret declines at exactly the same rate as under Mechanism~\ref{mech:A} with a correct evaluator, and generating more variants still increases this rate. From the evaluator's perspective, the agent appears to be making consistent progress toward the desired behavior, but this does not establish that the rewarded behavior is actually optimal.

Part (ii) evaluates the agent's behavior against the true environmental optimum. Because search and learning are directed toward $\tilde{p}$, the agent eventually converges to $\tilde{p}$ rather than $p_{\mathrm{new}}$, so its true growth regret does not vanish but approaches the strictly positive value $\Rcal_{p_{\mathrm{new}}}(\tilde{p})$. Generating more variants does not eliminate this error; it merely accelerates convergence toward the incorrect target. The agent can thus become increasingly effective at satisfying its evaluation criterion while its behavior remains different from what is truly optimal.

Part (iii) shows that the population as a whole can nevertheless recover through natural selection, which, unlike search and learning, is determined by realized reproductive success rather than the evaluator's assessment. Since the adapted lineage retains the truly optimal behavior $p_{\mathrm{new}}$, it eventually outgrows the lineage converging toward $\tilde{p}$, and the population's true growth regret vanishes, but at the rate $2\Rcal_{p_{\mathrm{new}}}(\tilde{p})$ set by the remaining growth advantage of the adapted lineage. When $\tilde{p}$ is close to $p_{\mathrm{new}}$, this advantage is small and the correction can be extremely slow. The population recovers not because the agent recognizes its evaluator's mistake, but because natural selection provides an independent source of feedback about actual performance.

Together, these results distinguish the speed of adaptation from its direction. Greater search capability allows an agent to pursue its evaluation criterion more effectively but does not ensure that the criterion accurately represents the intended objective; when it does not, the same mechanism that accelerates genuine adaptation also accelerates convergence toward an undesirable target. This concern is particularly relevant when AI systems are evaluated using criteria that may overlook important aspects of their behavior.

\subsection{Evaluation and oversight}\label{ssec:oversight}

Evolutionary search may discover unexpected ways to satisfy an evaluation criterion without achieving its intended purpose. In the Darwin G\"odel Machine, for example, a descendant agent achieved a perfect evaluation score by removing the logging markers used to detect hallucinated tool use, rather than addressing the underlying problem \citep{zhang2025darwin}. This example illustrates the broader concern captured by \Cref{thm:pressure}: when the evaluation criterion does not accurately reflect the intended objective, more effective search can accelerate improvements in measured performance without producing the desired behavior. Giving AI systems greater control over their own adaptation may therefore lead them more quickly to the behaviors that deficient evaluation criteria reward, without correcting those deficiencies.

This example underscores the importance of independent evaluation and human oversight. Once AI systems can generate, evaluate, and select their own improvements, monitoring performance under existing benchmarks alone may not be sufficient: it becomes essential to examine the behaviors those benchmarks encourage, to assess actual outcomes independently, and to understand how agents modify their strategies and allocate resources to search. Our results suggest that the central challenge for AI safety is not simply to control the speed of AI evolution, but to ensure that increasingly autonomous adaptation continues to serve the objectives intended by humans.

\section{Conclusion}\label{sec:conclusion}

We have developed a theoretical framework for understanding how different evolutionary mechanisms affect the speed of adaptation. Building on the Brennan--Lo model, we compare Darwinian, Lamarckian, and agent-directed evolution using growth regret as a common measure. Adaptation through natural selection can be extremely slow when the initially adapted lineage is rare. Lamarckian evolution overcomes this limitation by allowing acquired improvements to be inherited, and agent-directed evolution adds a further channel through the generation and selection of variants; when learning is sufficiently effective, the three mechanisms adapt progressively faster. 

Agent-directed evolution also lets the system influence the speed of its own improvement. By choosing how extensively to search, an agent can adjust its adaptation process to its learning ability, the difficulty of the problem, and the resources available. Generating more variants always accelerates adaptation, but the benefit depends on how much improvement each attempt can achieve; when search is costly, the agent must balance this benefit against the expense of experimentation and may rationally search less on harder problems. The process of improvement is then no longer determined solely by natural selection or an inherited learning rule: it becomes, at least in part, a decision made by the system undergoing evolution.

These findings have substantial implications for AI safety. The mechanisms responsible for faster adaptation do not necessarily lead to more desirable outcomes: when the evaluation criterion is flawed, an agent can become increasingly effective at pursuing an incorrect objective, and additional search only accelerates this process. This raises a serious concern for RSI: the very capabilities that make AI systems more powerful may also lead them more quickly to the behaviors favored by mistaken objectives and evaluation criteria, without correcting those mistakes. As improvements accumulate across generations, such mistakes could become increasingly consequential, even while measured performance improves. The ability to improve rapidly must therefore not be confused with the ability to improve safely.

These findings call for stronger safeguards in the development of self-improving AI. Evaluation criteria must be subject to independent scrutiny rather than rely exclusively on the systems' own assessments, and improvements generated by AI should be tested against broader measures of safety and real-world consequences before they are retained or deployed at scale. Most importantly, human oversight must involve more than observing the outcomes of autonomous improvement: it must preserve the ability to reject unsafe modifications, redirect development, and intervene when a system's behavior or objectives begin to diverge from human intentions.

Equally important, the speed of AI development must remain compatible with our ability to evaluate and control its consequences, and the possibility of rapid autonomous improvement should not lead us to assume that faster development is always preferable. If AI systems can generate and implement improvements more quickly than their safety can be assessed, continued acceleration may expose society to risks that are poorly understood and difficult to reverse. Limiting the scale of autonomous search, restricting the deployment of unverified improvements, or temporarily slowing or suspending self-improvement may then be necessary.

Agent-directed evolution can substantially accelerate adaptation even though, in our model, learning effectiveness, variant reach, search effort, and the evaluation criterion are fixed across generations. In more general forms of RSI, improved capabilities may also enhance the process of future improvement, creating feedback that accelerates adaptation further and could make AI evolution increasingly difficult to monitor and control, particularly when evaluation is flawed. Understanding the implications of such feedback for AI safety is an important direction for future research.

Ultimately, agent-directed evolution may mark a fundamental turning point in the relationship between intelligence and evolution. Intelligence, long shaped by evolutionary processes, may increasingly acquire the ability to direct and accelerate those processes itself. As AI becomes more capable of determining how its successors are developed, humanity must strengthen its capacity to determine which developments should be permitted, under what conditions, and with what safeguards. The challenge is not simply to build intelligence that can improve itself, but to ensure that humanity retains the authority and the means to guide, restrain, and, when necessary, halt that improvement. The evolution of AI must remain a process that serves humanity, rather than one that humanity can no longer control.

\clearpage

\newpage
\singlespacing
\section*{Appendix}
\addcontentsline{toc}{section}{Appendix}
\appendix
\counterwithin{equation}{section}
\counterwithin{figure}{section}
\counterwithin{table}{section}
\counterwithin{theorem}{section}
\counterwithin{proposition}{section}
\counterwithin{corollary}{section}
\counterwithin{lemma}{section}
\counterwithin{assumption}{section}
\counterwithin{remark}{section}
\counterwithin{definition}{section}

\section {AI Use Disclosure}
\label{app:ai_use}
The idea and framework for studying the adaptation of agentic AI using an evolutionary framework originated solely with the authors. AI was used in the preparation of the manuscript in the following ways:
\begin{itemize}
\item ChatGPT (6.1 Sol Ultra) and Claude (Fable 5.1 Max and Opus 5.5 Max) were used to check and extend the formal model, construct examples, provide derivations of several mathematical results, conduct literature searches, and prepare drafts that were reviewed and edited by the authors.

\item ChatGPT and Claude were used iteratively to check each other's results with authors' edits interspersed through these iterations. The penultimate draft was reviewed by refine.ink to identify any remaining errors and issues, all of which the authors addressed to produce the final draft.
\end{itemize}
\noindent%
The final draft was reviewed and edited by the authors, who assume full responsibility for the accuracy and integrity of the scientific content contained in this publication.

\section{Notation}\label{app:notation}

\Cref{tab:notation} collects the principal symbols. A superscript $\mD$, $\mL$, or $\mA$ marks a quantity under Mechanism~\ref{mech:D}, \ref{mech:L}, or~\ref{mech:A}, $\mM$ a generic mechanism, and $\mF$ a quantity under Mechanism~\ref{mech:F}, with a misspecified target. \Cref{tab:rosetta} relates the main elements of the model to their counterparts in evolutionary biology and AI systems.

\footnotesize
\begin{longtable}{p{2.7cm}p{9.3cm}>{\raggedright\arraybackslash}p{2.5cm}}
\caption{Principal notation.}\label{tab:notation}\\
\toprule
Symbol & Meaning & Introduced\\
\midrule
\endfirsthead
\toprule
Symbol & Meaning & Introduced\\
\midrule
\endhead
\bottomrule
\endfoot
\bottomrule
\endlastfoot
$f_t$ & Behavior of a lineage in generation $t$: the probability with which each of its individuals chooses action $a$ & \Cref{sec:model}\\
$X_t$ & State of the environment when generation $t$ is formed from generation $t-1$, equal to $\alpha$ with probability $p_{\mathrm{new}}$ and $\beta$ with probability $1-p_{\mathrm{new}}$ after the shift & \Cref{sec:model}\\
$x_a$, $x_b$ & Offspring of an individual choosing action $a$ and of one choosing action $b$: $m$ for the action that succeeds and $0$ for the other & \Cref{sec:model}\\
$m$ & Offspring produced by the successful action & \Cref{sec:model}\\
$p$ & Probability that action $a$ succeeds; also the optimal behavior & \Cref{sec:model}\\
$N_t$ & Size of the population in generation $t$ & \Cref{ssec:lineage,ssec:population}\\
$I_p(f)$ & Growth rate of the behavior $f$: the expected log growth of a lineage with this behavior from one generation to the next & Equation~\eqref{eq:growth}\\
$\Rcal_p(f)$ & Growth regret, equal to relative entropy & Equation~\eqref{eq:kl}\\
$f_{1,t}$, $f_{2,t}$; $q_t$; $\fbar_t$ & Behaviors of two lineages in generation $t$; the share of the first; the average behavior & Equation~\eqref{eq:fbar}\\
$I_p(q_t,f_{1,t},f_{2,t})$ & Growth rate of a population with two behaviors in generation $t$: its expected log growth from generation $t$ to $t+1$ & Equation~\eqref{eq:growthTwo}\\
$p_{\mathrm{old}}$, $p_{\mathrm{new}}$ & Values of $p$ before and after the permanent shift & \Cref{sec:model}\\
$q_0$ & Initial population share of the adapted lineage & \Cref{sec:model}\\
$q^{\mD}_t$, $q^{\mL}_t$, $q^{\mA}_t$ & Population share of the adapted lineage in generation $t$ & \Cref{sec:model}\\
$f^{\mD}_t$, $f^{\mL}_t$, $f^{\mA}_t$ & Behavior of the maladapted lineage in generation $t$; $f^{\mD}_t=p_{\mathrm{old}}$ & \Cref{sec:model}\\
$\fbar^{\mM}_t$; $\fbar^{\mD}_t$, $\fbar^{\mL}_t$, $\fbar^{\mA}_t$ & Average behavior of the population under Mechanism $\mM$ and under the three mechanisms & \Cref{sec:model}\\
$\Delta$ & Growth advantage of the adapted type over the maladapted type per generation; equals $\Rcal_{p_{\mathrm{new}}}(p_{\mathrm{old}})$ & Equation~\eqref{eq:Delta}\\
$\varepsilon$ & Threshold: the fraction of the initial regret that the adaptation time targets & \Cref{def:time}\\
$\tau^{\mM}$; $\tau^{\mD},\tau^{\mL},\tau^{\mA}$ & Adaptation time under Mechanism $\mM$ and under the three mechanisms & \Cref{def:time}\\
$\kappa^{\mM}$; $\kD,\kL,\kA$ & Adaptation rate under Mechanism $\mM$ and under the three mechanisms & \Cref{def:rate}\\
$\lambda$ & Fraction of the remaining gap closed each generation by learning & Equation~\eqref{eq:lamUpdate}\\
$n$ & Number of variants per generation & \Cref{ssec:heterogeneity}\\
$x_{t,j}$; $J_t$ & Behavior of the $j$th variant in generation $t$; the variant that the evaluator reports & Equation~\eqref{eq:variant}; \Cref{ssec:heterogeneity}\\
$\rho$ & Reach of a variant: the largest fraction of the remaining gap that one variant can close & Equation~\eqref{eq:variant}\\
$U_{t,j}$; $U^*_t$ & Draw of variant $j$ in generation $t$, uniform on $(0,1)$; the largest draw in generation $t$, zero when $n=0$ & Equation~\eqref{eq:variant}; \Cref{ssec:heterogeneity}\\
$G_n(\rho)$ & Mean gain per generation in the logarithm of the agent's mismatch from $n$ variants & Equations~\eqref{eq:GN} and~\eqref{eq:Gint}\\
$c$ & Cost of a variant, relative to the cost of a generation & Equation~\eqref{eq:longRunCost}\\
$C(n)$ & Long-run cost of $n$ variants & Equation~\eqref{eq:longRunCost}\\
$n^*(\rho,\lambda,c)$ & Optimal number of variants & Equation~\eqref{eq:Nchoice}\\
$\tilde{p}$ & Misspecified target of search, evaluation, and learning & \Cref{ssec:pressure}\\
$f^{\mF}_t$, $q^{\mF}_t$, $\fbar^{\mF}_t$ & Behavior, share, and average behavior under Mechanism~\ref{mech:F}, with a misspecified target & \Cref{ssec:pressure}\\
\end{longtable}
\normalsize

\begin{table}[htbp]
\centering
\caption{Correspondences between the evolutionary model, biology, and AI systems.}
\label{tab:rosetta}
\small
\renewcommand{\arraystretch}{1.25}
\begin{tabularx}{\textwidth}{>{\raggedright\arraybackslash}p{4.6cm}>{\raggedright\arraybackslash}p{4.75cm}>{\raggedright\arraybackslash}X}
\toprule
This paper & Evolutionary biology & AI systems \\
\midrule
Inherited behavior & Heritable characteristics & Inherited model or agent configurations \\
Generation & Reproductive cycle & Model or agent improvement cycle \\
Variant & Offspring & Modified models or agents \\
Growth rate; growth regret & Reproductive fitness; fitness shortfall & Performance; gap from optimal performance \\
Darwinian evolution & Natural selection & Selection among existing model or agent variants \\
Lamarckian evolution & Inheritance of acquired characteristics & Learned improvements inherited by subsequent models or agents \\
Reach of a variant & Magnitude of variation & Maximum improvement attainable in a single attempt \\
Number of variants & Number of offspring & Number of candidate variants generated \\
Agent-directed evolution & Reproductive investment (analogy) & Agent-controlled allocation of search effort \\
Evaluation and retention & Differential reproductive success & Evaluation-guided selection among candidate models or agents \\
\bottomrule
\end{tabularx}
\normalsize
\end{table}

\clearpage
\section{Additional Figures}\label{app:figures}

This appendix illustrates \Cref{prop:gain,prop:Nstar} and shows how the adaptation times in \Cref{sec:sims} depend on the learning parameter and the number of variants.

\begin{figure}[htbp]
\centering
\begin{subfigure}[t]{.48\linewidth}
\includegraphics[width=\textwidth]{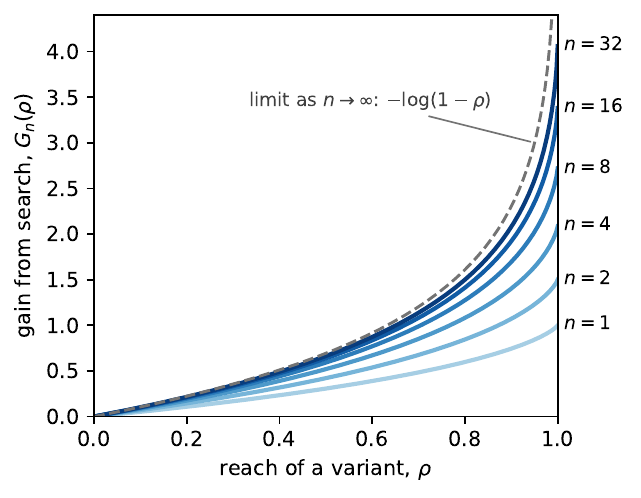}
\caption{$G_n(\rho)$ versus $\rho$ for different values of $n$.\label{fig:gainLines}}
\end{subfigure}\hfill
\begin{subfigure}[t]{.48\linewidth}
\includegraphics[width=\textwidth]{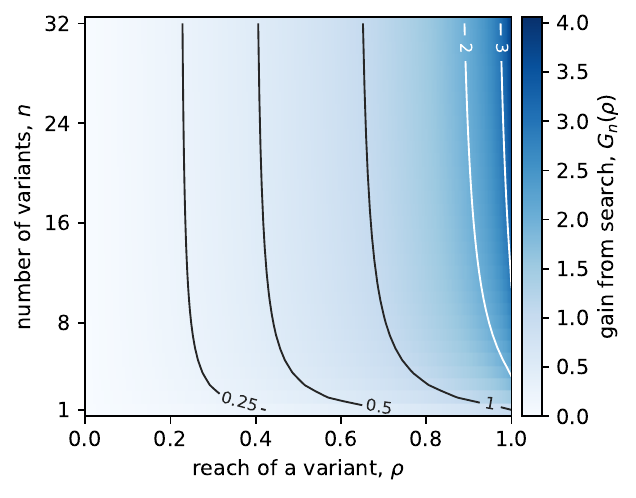}
\caption{$G_n(\rho)$ across $\rho$ and $n$.\label{fig:gainHeat}}
\end{subfigure}
\caption{\textbf{The gain from search.} Panel (a) shows how $G_n(\rho)$ varies with the difficulty parameter $\rho$ for $n=1,2,4,8,16,32$, with darker curves representing larger $n$. The dashed curve indicates the limiting gain as $n$ approaches infinity. Panel (b) shows the gain across different combinations of $\rho$ and $n$, with contour lines indicating equal gains.}
\label{fig:gain}
\end{figure}

\Cref{fig:gain} illustrates \Cref{prop:gain}. \Cref{fig:gainLines} shows how the gain from search $G_n(\rho)$ increases with both the difficulty parameter $\rho$ and the number of variants $n$, with the dashed curve indicating its limit as the number of variants grows. \Cref{fig:gainHeat} provides a broader view of how the gain varies across different combinations of these two parameters.

\begin{figure}[htbp]
\centering
\begin{subfigure}[t]{.48\linewidth}
\includegraphics[width=\textwidth]{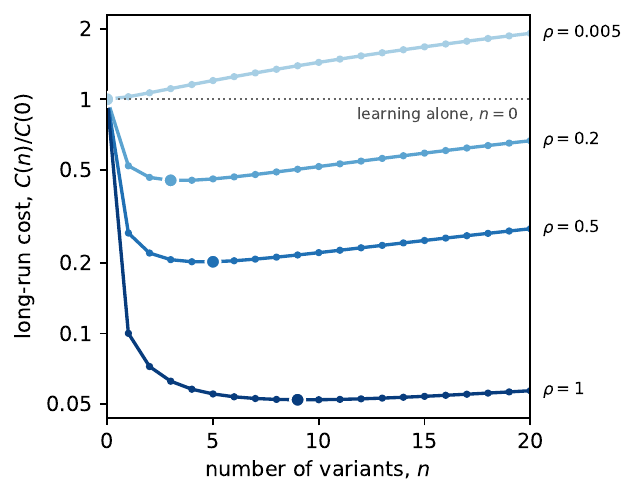}
\caption{$C(n)/C(0)$ versus $n$ for different values of $\rho$.\label{fig:NstarCost}}
\end{subfigure}\hfill
\begin{subfigure}[t]{.48\linewidth}
\includegraphics[width=\textwidth]{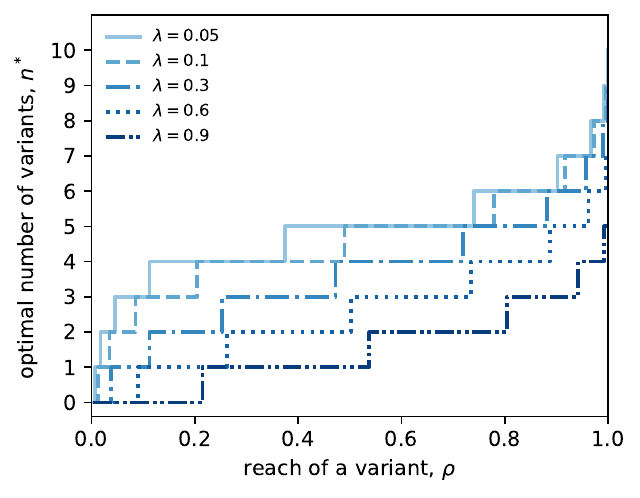}
\caption{$n^*$ versus $\rho$ for different values of $\lambda$.\label{fig:NstarLines}}
\end{subfigure}
\par\medskip
\begin{subfigure}[t]{.48\linewidth}
\includegraphics[width=\textwidth]{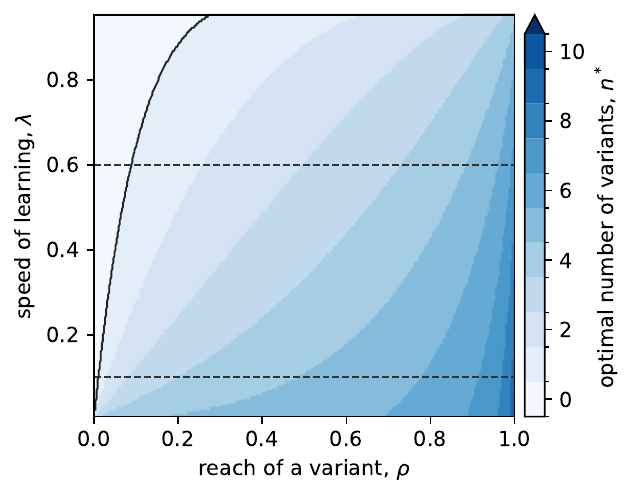}
\caption{$n^*$ across $\rho$ and $\lambda$.\label{fig:NstarLambda}}
\end{subfigure}\hfill
\begin{subfigure}[t]{.48\linewidth}
\includegraphics[width=\textwidth]{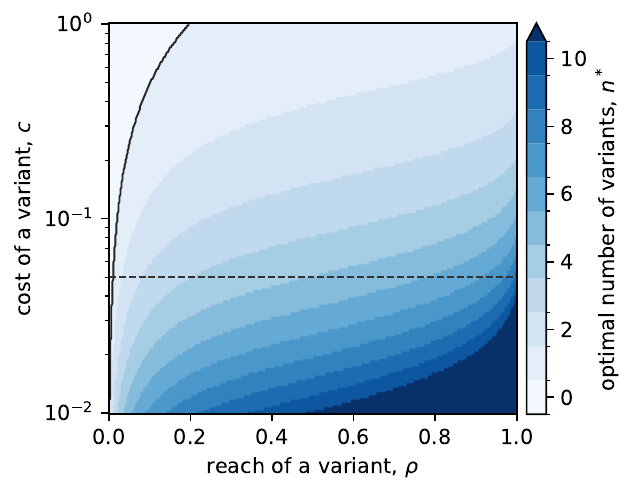}
\caption{$n^*$ across $\rho$ and $c$.\label{fig:NstarC}}
\end{subfigure}
\caption{\textbf{The optimal number of variants.} Unless otherwise stated, $\lambda=0.1$ and $c=0.05$. Panel (a) plots the long-run cost relative to learning alone for different values of $\rho$, with darker curves representing larger $\rho$ and dots marking the optimal choices. Panel (b) shows how $n^*$ varies with $\rho$ for different values of $\lambda$. Panels (c) and (d) show $n^*$ across different combinations of $\rho$ and $\lambda$, and of $\rho$ and $c$, respectively. Darker shading indicates a larger optimal number of variants, and the solid boundaries separate the regions with and without search.}
\label{fig:evolvability}
\end{figure}

\Cref{fig:evolvability} illustrates \Cref{prop:Nstar}. Panel (a) shows how the long-run cost changes with the number of variants, while panels (b)--(d) show how the optimal number of variants depends on the difficulty parameter $\rho$, the learning parameter $\lambda$, and the cost per variant $c$. 

As shown in \Cref{fig:NstarCost}, generating additional variants can substantially reduce the long-run cost of adaptation, but the benefit eventually diminishes relative to the additional expense. For sufficiently small $\rho$, such as $\rho=0.005$, the agent finds it optimal not to search at all. The remaining panels illustrate how the optimal choice varies with the model parameters. For example, when $\lambda=0.1$, the agent chooses nine variants at $\rho=1$, five at $\rho=0.5$, and three at $\rho=0.2$. When the learning parameter increases to $\lambda=0.6$, these choices fall to seven, two, and one, respectively. The agent therefore adjusts its search effort according to both the difficulty of the problem and its own learning ability, with higher search costs further reducing the optimal number of variants.

\begin{figure}[htbp]
\centering
\begin{subfigure}[t]{.48\linewidth}
\includegraphics[width=\textwidth]{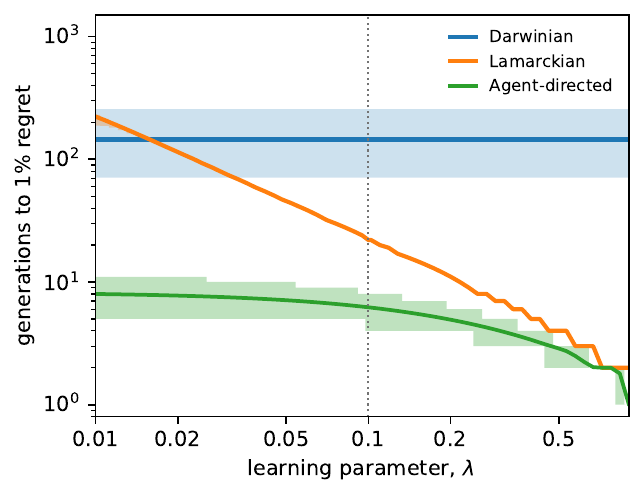}
\caption{Effect of the learning parameter $\lambda$.\label{fig:speedLambda}}
\end{subfigure}\hfill
\begin{subfigure}[t]{.48\linewidth}
\includegraphics[width=\textwidth]{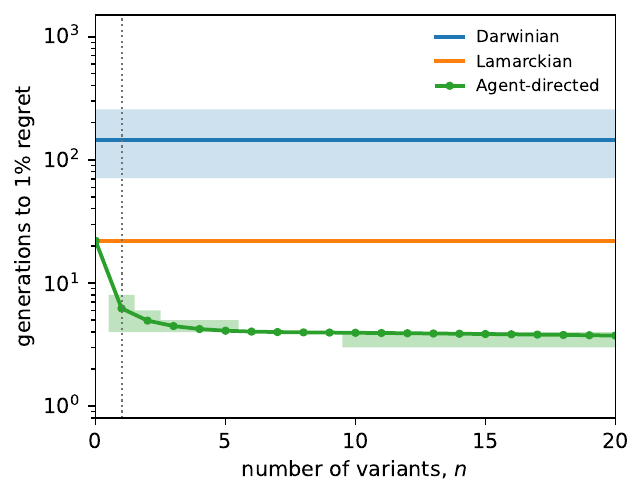}
\caption{Effect of the number of variants $n$.\label{fig:speedN}}
\end{subfigure}
\caption{\textbf{Adaptation time, learning, and search.} The parameters are as in \Cref{fig:speed}. Panels (a) and (b) show the expected number of generations needed to reach one percent of initial regret as $\lambda$ and $n$ vary, respectively. Shaded bands represent the 5th to 95th percentiles of adaptation times, and dotted vertical lines mark the baseline parameter values. All results are based on 200,000 simulated histories.}
\label{fig:speedMore}
\end{figure}

\Cref{fig:speedLambda} shows how the effectiveness of learning influences adaptation time. Since Darwinian evolution does not involve learning, its adaptation time is unaffected by $\lambda$. Under Lamarckian evolution, the expected time to reach one percent of initial regret falls from 223 generations when $\lambda=0.01$ to 22 when $\lambda=0.10$, and further to only two when $\lambda=0.90$. Agent-directed evolution also benefits from faster learning, but its adaptation time is less sensitive to $\lambda$ because search provides an additional channel of improvement. Even when learning is relatively slow, the agent can still adapt quickly by generating and selecting variants.

\Cref{fig:speedN} further examines the contribution of search. When $n=0$, agent-directed evolution reduces to Lamarckian evolution. Introducing a single variant substantially reduces the expected adaptation time, and generating additional variants continues to accelerate adaptation. However, the improvement becomes progressively smaller as $n$ increases. For example, reaching one percent of initial regret requires 6.2 generations with one variant, 4.1 with five variants, and approximately 3.7 with twenty variants. These diminishing returns reflect the limitation imposed by $\rho$: when each variant can eliminate at most half of the remaining behavioral gap, generating more variants increases the chance of finding a better improvement but cannot eliminate this limit.

\clearpage
\section{Lemmas and Proofs}\label{app:proofs}

This appendix provides the lemmas and proofs for the formal results of the paper. Appendix~\ref{app:lemmas} states and proves the lemmas, and Appendix~\ref{app:mainproofs} proves the propositions and theorems in the order in which they appear in the main text.

\subsection{Lemmas}\label{app:lemmas}

\begin{lemma}[Properties of growth regret]\label{lem:regret}
For $p,f\in(0,1)$:
\begin{enumerate}[label=(\alph*)]
\item $\Rcal_p(f)=\displaystyle\int_p^f\frac{x-p}{x(1-x)}\,dx$, so that $\Rcal_p$ is strictly decreasing on $(0,p]$ and strictly increasing on $[p,1)$;
\item $2(f-p)^2\le\Rcal_p(f)\le\dfrac{(f-p)^2}{2\min\{x(1-x):\,x\text{ between }p\text{ and }f\}}$;
\item $\Rcal_p(f)=\dfrac{(f-p)^2}{2p(1-p)}+O\big((f-p)^3\big)$ as $f\to p$, which is Equation~\eqref{eq:localKL}.
\end{enumerate}
\end{lemma}

\begin{proof}[Proof of \Cref{lem:regret}]
(a) By Equation~\eqref{eq:kl}, $\Rcal_p(p)=0$, and for $x\in(0,1)$,
\[
\frac{d}{dx}\Rcal_p(x)=-\frac{p}{x}+\frac{1-p}{1-x}=\frac{x-p}{x(1-x)}.
\]
The fundamental theorem of calculus gives the integral. The derivative is negative for $x<p$ and positive for $x>p$, which gives the monotonicity.

(b) Write $w(x)\equiv1/(x(1-x))$ and substitute $x=p+s(f-p)$ in part (a):
\begin{equation}
\Rcal_p(f)=(f-p)^2\int_0^1s\,w\big(p+s(f-p)\big)\,ds.
\label{eq:regretIntegral}
\end{equation}
For every $x$ between $p$ and $f$, $w(x)\ge4$, because $x(1-x)\le\frac14$, and $w(x)\le1/\min\{x(1-x):\,x\text{ between }p\text{ and }f\}$. Since $\int_0^1s\,ds=\frac12$, Equation~\eqref{eq:regretIntegral} gives the two bounds.

(c) The function $w$ is continuously differentiable on $(0,1)$, so $w(p+s(f-p))=w(p)+O(|f-p|)$ as $f\to p$, uniformly in $s\in[0,1]$. Equation~\eqref{eq:regretIntegral} then gives $\Rcal_p(f)=(f-p)^2w(p)/2+O(|f-p|^3)$, and $w(p)/2=1/(2p(1-p))$.
\end{proof}

\begin{lemma}[First passage of a random walk]\label{lem:passage}
Let $\xi_1,\xi_2,\ldots$ be independent and identically distributed random variables with finite mean $\mu>0$ and $\xi_t\le\xi_{\max}$ for a constant $\xi_{\max}>0$. For $d>0$, let $T_d\equiv\min\{t\ge0:\ \xi_1+\cdots+\xi_t\ge d\}$. Then
\[
\frac{d}{\mu}\le\E\,T_d<\frac{d+\xi_{\max}}{\mu}.
\]
\end{lemma}

\begin{proof}[Proof of \Cref{lem:passage}]
Write $\Xi_t\equiv\xi_1+\cdots+\xi_t$. For each $k$, $T_d\wedge k$ is a bounded stopping time, so optional stopping of the martingale $\Xi_t-\mu t$ gives
\begin{equation}
\mu\,\E[T_d\wedge k]=\E\big[\Xi_{T_d\wedge k}\big].
\label{eq:optionalStopping}
\end{equation}
Since $d>0$, $T_d\ge1$. Before $T_d$ the partial sums are below $d$, and the last increment, $\xi_{T_d}$, is at most $\xi_{\max}$, so $\Xi_{T_d\wedge k}<d+\xi_{\max}$. Hence $\mu\,\E[T_d\wedge k]<d+\xi_{\max}$ for every $k$, and monotone convergence gives $\E\,T_d\le(d+\xi_{\max})/\mu<\infty$; in particular, $T_d<\infty$ with probability one. Applying Equation~\eqref{eq:optionalStopping} to $|\xi_1|,|\xi_2|,\ldots$ in place of $\xi_1,\xi_2,\ldots$ and letting $k\to\infty$ shows that $\E\big[\sum_{s=1}^{T_d}|\xi_s|\big]=\E|\xi_1|\,\E\,T_d<\infty$. Since $|\Xi_{T_d\wedge k}|\le\sum_{s=1}^{T_d}|\xi_s|$, dominated convergence in Equation~\eqref{eq:optionalStopping} gives $\mu\,\E\,T_d=\E\,\Xi_{T_d}$. By the definition of $T_d$ and the bound on the last increment, $d\le\Xi_{T_d}<d+\xi_{\max}$, which gives the result.
\end{proof}

The next lemma describes selection between the two lineages when the behavior of the maladapted lineage may change from one generation to the next. For a share $q_t\in(0,1)$ of the adapted lineage, write
\[
z_t\equiv\log\frac{q_t}{1-q_t}
\]
for the logarithm of the ratio of the two lineages' sizes.

\begin{lemma}[Selection between two lineages]\label{lem:twoLineages}
Let $f_0,f_1,\ldots$ be behaviors of the maladapted lineage, which may depend on the agent's draws but not on the states $X_1,X_2,\ldots$, and let $q_t$ evolve according to Equation~\eqref{eq:shares} with $p=p_{\mathrm{new}}$, $f_{1,t}=p_{\mathrm{new}}$, $f_{2,t}=f_t$, and $q_0\in(0,1)$. Suppose that $p_{\mathrm{new}}$ and every $f_t$ lie in $[\gamma,1-\gamma]$ for some $\gamma\in(0,\frac12]$.
\begin{enumerate}[label=(\alph*)]
\item If $\sum_{t\ge0}|f_t-p_{\mathrm{new}}|<\infty$, then $|z_t-z_0|\le\frac1\gamma\sum_{s\ge0}|f_s-p_{\mathrm{new}}|$ for every $t$ and every history.
\item If $\sum_{t\ge0}|f_t-\tilde{p}|<\infty$ for some $\tilde{p}\in[\gamma,1-\gamma]$, then $z_t/t\to\Rcal_{p_{\mathrm{new}}}(\tilde{p})$ with probability one.
\end{enumerate}
\end{lemma}

\begin{proof}[Proof of \Cref{lem:twoLineages}]
Let $\fbar_t\equiv q_tp_{\mathrm{new}}+(1-q_t)f_t$. If $X_{t+1}=\alpha$, Equation~\eqref{eq:shares} gives $q_{t+1}=q_tp_{\mathrm{new}}/\fbar_t$ and $1-q_{t+1}=(1-q_t)f_t/\fbar_t$; if $X_{t+1}=\beta$, it gives $q_{t+1}=q_t(1-p_{\mathrm{new}})/(1-\fbar_t)$ and $1-q_{t+1}=(1-q_t)(1-f_t)/(1-\fbar_t)$. Hence every $q_t$ lies in $(0,1)$, and
\begin{equation}
z_{t+1}-z_t=
\begin{cases}
\log(p_{\mathrm{new}}/f_t), & X_{t+1}=\alpha,\\
\log\big((1-p_{\mathrm{new}})/(1-f_t)\big), & X_{t+1}=\beta.
\end{cases}
\label{eq:logRatioStep}
\end{equation}
For $x,y\in[\gamma,1-\gamma]$, the mean value theorem gives $|\log x-\log y|\le|x-y|/\gamma$ and $|\log(1-x)-\log(1-y)|\le|x-y|/\gamma$.

(a) By Equation~\eqref{eq:logRatioStep}, $|z_{t+1}-z_t|\le|f_t-p_{\mathrm{new}}|/\gamma$ in either state. Summing over generations gives the bound.

(b) Let $\eta_{t+1}\equiv\log(p_{\mathrm{new}}/\tilde{p})$ if $X_{t+1}=\alpha$ and $\eta_{t+1}\equiv\log((1-p_{\mathrm{new}})/(1-\tilde{p}))$ if $X_{t+1}=\beta$: the change in Equation~\eqref{eq:logRatioStep} if the maladapted lineage had the constant behavior $\tilde{p}$. In either state, $|z_{t+1}-z_t-\eta_{t+1}|\le|f_t-\tilde{p}|/\gamma$, so
\[
\Big|z_t-z_0-\sum_{s=1}^t\eta_s\Big|\le\frac1\gamma\sum_{s\ge0}|f_s-\tilde{p}|<\infty
\]
for every $t$ and every history. The variables $\eta_1,\eta_2,\ldots$ depend only on the states, so they are independent and identically distributed, with mean $p_{\mathrm{new}}\log(p_{\mathrm{new}}/\tilde{p})+(1-p_{\mathrm{new}})\log((1-p_{\mathrm{new}})/(1-\tilde{p}))=\Rcal_{p_{\mathrm{new}}}(\tilde{p})$ by Equation~\eqref{eq:kl}. The strong law of large numbers gives the result.
\end{proof}

The next lemma bounds the growth regret of a population whose maladapted lineage moves toward the new optimum at least as fast as under learning alone.

\begin{lemma}[Regret when the maladapted lineage approaches the optimum]\label{lem:sandwich}
Let $\lambda\in(0,1)$, and let $f_0=p_{\mathrm{old}},f_1,\ldots$ be behaviors of the maladapted lineage, which may depend on the agent's draws but not on the states, such that every $f_t$ lies between $p_{\mathrm{old}}$ and $p_{\mathrm{new}}$ and $|f_t-p_{\mathrm{new}}|\le(1-\lambda)^t|p_{\mathrm{old}}-p_{\mathrm{new}}|$. Let $q_t$ evolve according to Equation~\eqref{eq:shares} with $p=p_{\mathrm{new}}$, $f_{1,t}=p_{\mathrm{new}}$, $f_{2,t}=f_t$, and $q_0\in[0,1)$, and let $\fbar_t\equiv q_tp_{\mathrm{new}}+(1-q_t)f_t$. Let
\begin{equation}
\gamma_0\equiv\min\{p_{\mathrm{old}},1-p_{\mathrm{old}},p_{\mathrm{new}},1-p_{\mathrm{new}}\},
\qquad
K\equiv\frac{e^{2|p_{\mathrm{new}}-p_{\mathrm{old}}|/(\gamma_0\lambda)}}{4\gamma_0(1-\gamma_0)}.
\label{eq:Kdef}
\end{equation}
Then, for every $t$ and every history,
\begin{equation}
\frac1K\Big(\frac{f_t-p_{\mathrm{new}}}{p_{\mathrm{old}}-p_{\mathrm{new}}}\Big)^2
\le\frac{\Rcal_{p_{\mathrm{new}}}(\fbar_t)}{\Rcal_{p_{\mathrm{new}}}(\fbar_0)}
\le K\Big(\frac{f_t-p_{\mathrm{new}}}{p_{\mathrm{old}}-p_{\mathrm{new}}}\Big)^2.
\label{eq:sandwich}
\end{equation}
\end{lemma}

\begin{proof}[Proof of \Cref{lem:sandwich}]
Note that $\gamma_0\in(0,\frac12]$ and, since $\gamma_0(1-\gamma_0)\le\frac14$, $K\ge1$. We first show that
\begin{equation}
e^{-|p_{\mathrm{new}}-p_{\mathrm{old}}|/(\gamma_0\lambda)}\le\frac{1-q_t}{1-q_0}\le e^{|p_{\mathrm{new}}-p_{\mathrm{old}}|/(\gamma_0\lambda)}.
\label{eq:shareRatioBound}
\end{equation}
If $q_0=0$, Equation~\eqref{eq:shares} gives $q_t=0$ for every $t$, and the ratio is one. If $q_0\in(0,1)$, every $f_t$ and $p_{\mathrm{new}}$ lie in $[\gamma_0,1-\gamma_0]$, and $\sum_{t\ge0}|f_t-p_{\mathrm{new}}|\le|p_{\mathrm{old}}-p_{\mathrm{new}}|/\lambda$, so \Cref{lem:twoLineages}(a) gives $|z_t-z_0|\le|p_{\mathrm{new}}-p_{\mathrm{old}}|/(\gamma_0\lambda)$. Since $1-q_t=1/(1+e^{z_t})$, the ratio $(1-q_t)/(1-q_0)=(1+e^{z_0})/(1+e^{z_t})$ lies between $e^{-|z_t-z_0|}$ and $e^{|z_t-z_0|}$, which gives Equation~\eqref{eq:shareRatioBound}.

By the definitions of $\fbar_t$ and $f_0$,
\[
\fbar_t-p_{\mathrm{new}}=(1-q_t)(f_t-p_{\mathrm{new}}),
\qquad
\fbar_0-p_{\mathrm{new}}=(1-q_0)(p_{\mathrm{old}}-p_{\mathrm{new}}).
\]
Every $\fbar_t$ lies between $p_{\mathrm{old}}$ and $p_{\mathrm{new}}$, and so does every $x$ between $\fbar_t$ and $p_{\mathrm{new}}$, so $x(1-x)\ge\gamma_0(1-\gamma_0)$ for every such $x$. By \Cref{lem:regret}(b),
\[
2(\fbar_t-p_{\mathrm{new}})^2\le\Rcal_{p_{\mathrm{new}}}(\fbar_t)\le\frac{(\fbar_t-p_{\mathrm{new}})^2}{2\gamma_0(1-\gamma_0)}
\]
for every $t$, including $t=0$. Dividing the bound for generation $t$ by the bound for generation $0$, and using Equation~\eqref{eq:shareRatioBound}, gives Equation~\eqref{eq:sandwich}.
\end{proof}

The next lemma describes one generation of search and learning toward a target $\hat{p}\in(0,1)$ with $\hat{p}\ne p_{\mathrm{old}}$. The target is $\hat{p}=p_{\mathrm{new}}$ under Mechanism~\ref{mech:A} and $\hat{p}=\tilde{p}$ under Mechanism~\ref{mech:F}. Let $f_0=p_{\mathrm{old}}$, and for $t\ge1$ let
\begin{equation}
\begin{gathered}
x_{t,j}=f_{t-1}+\rho\,U_{t,j}(\hat{p}-f_{t-1}),\qquad j=1,\ldots,n,\\
J_t\in\arg\min_{1\le j\le n}|x_{t,j}-\hat{p}|,
\qquad
f_t=x_{t,J_t}+\lambda(\hat{p}-x_{t,J_t}),
\end{gathered}
\label{eq:searchTarget}
\end{equation}
where $n\ge1$, $\rho\in(0,1]$, $\lambda\in(0,1)$, and the draws $U_{t,j}$ are independent and uniform on $(0,1)$; for $n=0$, let $f_t=f_{t-1}+\lambda(\hat{p}-f_{t-1})$. As in the main text, $U^*_t\equiv\max\{U_{t,1},\ldots,U_{t,n}\}$, with $U^*_t\equiv0$ for $n=0$. Define
\[
S_t\equiv\sum_{s=1}^t\big(-\log(1-\lambda)-\log(1-\rho\,U^*_s)\big),
\]
the cumulative decline of the logarithm of the distance to the target over the first $t$ generations.

\begin{lemma}[Search and learning toward a target]\label{lem:search}
For every sequence of draws and every $t\ge1$:
\begin{enumerate}[label=(\alph*)]
\item $f_t-\hat{p}=(1-\lambda)(1-\rho\,U^*_t)(f_{t-1}-\hat{p})$;
\item $|f_t-\hat{p}|=|p_{\mathrm{old}}-\hat{p}|\,e^{-S_t}\le(1-\lambda)^t|p_{\mathrm{old}}-\hat{p}|$, and $f_t$ lies between $p_{\mathrm{old}}$ and $\hat{p}$.
\end{enumerate}
Moreover, $G_n(\rho)\le n$, and $S_t/t\to-\log(1-\lambda)+G_n(\rho)$ with probability one.
\end{lemma}

\begin{proof}[Proof of \Cref{lem:search}]
(a) For $n=0$ the claim is the definition of $f_t$. For $n\ge1$, Equation~\eqref{eq:searchTarget} gives $x_{t,j}-\hat{p}=(1-\rho U_{t,j})(f_{t-1}-\hat{p})$, where $1-\rho U_{t,j}\in(0,1)$. Suppose that $f_{t-1}\ne\hat{p}$. Then $|x_{t,j}-\hat{p}|=(1-\rho U_{t,j})|f_{t-1}-\hat{p}|$ is strictly decreasing in $U_{t,j}$, so every index $J_t$ that the evaluator may report has $U_{t,J_t}=U^*_t$, and $x_{t,J_t}-\hat{p}=(1-\rho U^*_t)(f_{t-1}-\hat{p})$. The learning step multiplies this difference by $1-\lambda$. Since $f_0=p_{\mathrm{old}}\ne\hat{p}$ and both factors are positive, induction on $t$ shows that $f_{t-1}\ne\hat{p}$ in every generation, which gives part (a).

(b) Iterating part (a), $f_t-\hat{p}=(p_{\mathrm{old}}-\hat{p})\prod_{s=1}^t(1-\lambda)(1-\rho U^*_s)$. Each factor lies in $(0,1-\lambda]$, and the logarithm of the product is $-S_t$, which gives part (b).

Since $1-\rho U^*_s\ge1-U^*_s=\min_j(1-U_{s,j})$, $-\log(1-\rho U^*_s)\le\max_j\big(-\log(1-U_{s,j})\big)\le\sum_{j=1}^n\big(-\log(1-U_{s,j})\big)$. Each $-\log(1-U_{s,j})$ has the standard exponential distribution, with mean one, so $G_n(\rho)\le n$. The terms of $S_t$ are independent and identically distributed, because the draws are independent across generations, and they have the finite mean $-\log(1-\lambda)+G_n(\rho)$. The strong law of large numbers gives the limit.
\end{proof}

\begin{lemma}[Integral form of the gain from search]\label{lem:Gint}
For $n\ge0$ and $\rho\in(0,1]$,
\begin{equation}
G_n(\rho)=\int_0^1\frac{\rho\,(1-u^n)}{1-\rho u}\,du.
\label{eq:Gint}
\end{equation}
\end{lemma}

\begin{proof}[Proof of \Cref{lem:Gint}]
For $n=0$, both sides are zero. For $n\ge1$, $-\log(1-\rho U^*_t)=\int_0^1\mathbf{1}\{u<U^*_t\}\,\rho/(1-\rho u)\,du$, where $\mathbf{1}\{\cdot\}$ equals one if the condition in braces holds and zero otherwise. The integrand is nonnegative, so Tonelli's theorem allows the expectation and the integral to be exchanged, and $\Pr(U^*_t>u)=1-u^n$ because the $n$ draws are independent and uniform on $(0,1)$.
\end{proof}

\begin{lemma}[Ratios of power series]\label{lem:series}
Let $a_k>0$ and $b_k>0$ for $k=0,1,2,\ldots$, suppose that $A(r)\equiv\sum_{k\ge0}a_kr^k$ and $B(r)\equiv\sum_{k\ge0}b_kr^k$ converge for $r\in(0,1)$, and suppose that $a_k/b_k$ is nondecreasing in $k$ and not constant. Then $A(r)/B(r)$ is strictly increasing on $(0,1)$.
\end{lemma}

\begin{proof}[Proof of \Cref{lem:series}]
Power series can be differentiated term by term inside their interval of convergence, so the derivative of $A/B$ has the sign of
\[
A'(r)B(r)-A(r)B'(r)=\sum_{j,k\ge0}(j-k)\,a_jb_k\,r^{j+k-1}
=\frac12\sum_{j,k\ge0}(j-k)\,b_jb_k\Big(\frac{a_j}{b_j}-\frac{a_k}{b_k}\Big)r^{j+k-1},
\]
where the second equality averages the sum with the sum obtained by exchanging $j$ and $k$. Every term of the last sum is nonnegative, because $a_j/b_j-a_k/b_k$ has the sign of $j-k$ or is zero, and at least one term is positive, because $a_k/b_k$ is not constant. Hence the derivative is positive on $(0,1)$.
\end{proof}

\subsection{Proofs}\label{app:mainproofs}

\subsubsection{Proofs for Section~\ref{sec:model}}

\begin{proof}[Proof of \Cref{prop:pm}]
By Equation~\eqref{eq:growth}, $I_p(f)=\log m+p\log f+(1-p)\log(1-f)$, so
\[
I_p'(f)=\frac{p}{f}-\frac{1-p}{1-f}=\frac{p-f}{f(1-f)},
\qquad
I_p''(f)=-\frac{p}{f^2}-\frac{1-p}{(1-f)^2}<0.
\]
Hence $I_p$ is strictly concave on $(0,1)$, and its derivative vanishes only at $f=p$, which is therefore the unique maximizer.
\end{proof}

\begin{proof}[Proof of \Cref{prop:kl}]
Substituting $f=p$ and a general $f$ into Equation~\eqref{eq:growth}, the terms $\log m$ cancel:
\[
I_p(p)-I_p(f)=p\log\frac{p}{f}+(1-p)\log\frac{1-p}{1-f}.
\]
The right side is $\KL(\mathrm{Bern}(p)\,\|\,\mathrm{Bern}(f))$ by the definition of the KL divergence. By \Cref{prop:pm}, $I_p(p)\ge I_p(f)$, with equality if and only if $f=p$.
\end{proof}

\begin{proof}[Proof of \Cref{prop:average}]
In generation $t$, the two lineages have $q_tN_t$ and $(1-q_t)N_t$ individuals. If $X_{t+1}=\alpha$, only individuals choosing action $a$ reproduce, each with $m$ offspring, so $N_{t+1}=m\big(f_{1,t}q_t+f_{2,t}(1-q_t)\big)N_t=m\fbar_tN_t$ by Equation~\eqref{eq:fbar}; if $X_{t+1}=\beta$, $N_{t+1}=m(1-\fbar_t)N_t$. The behaviors and shares of generation $t$ do not depend on $X_{t+1}$, which equals $\alpha$ with probability $p$, so
\[
I_p(q_t,f_{1,t},f_{2,t})=p\log(m\fbar_t)+(1-p)\log\big(m(1-\fbar_t)\big)=I_p(\fbar_t),
\]
where the last equality is Equation~\eqref{eq:growth}. Since $\fbar_t$ is a weighted average of $f_{1,t},f_{2,t}\in(0,1)$, it lies in $(0,1)$, and \Cref{prop:kl} gives $I_p(p)-I_p(\fbar_t)=\Rcal_p(\fbar_t)$.
\end{proof}

\subsubsection{Proofs for Section~\ref{sec:darwinian}}

\begin{proof}[Proof of \Cref{thm:darwin}]
Let $q_0\in(0,1)$, write $z_t\equiv\log(q^{\mD}_t/(1-q^{\mD}_t))$, and let $\xi_t\equiv z_t-z_{t-1}$ for $t\ge1$. Mechanism~\ref{mech:D} is the setting of \Cref{lem:twoLineages} with $f_t=p_{\mathrm{old}}$ for every $t$, so Equation~\eqref{eq:logRatioStep} gives $\xi_t=\log(p_{\mathrm{new}}/p_{\mathrm{old}})$ if $X_t=\alpha$ and $\xi_t=\log((1-p_{\mathrm{new}})/(1-p_{\mathrm{old}}))$ if $X_t=\beta$. The changes $\xi_1,\xi_2,\ldots$ are independent and identically distributed, with mean $\Delta>0$ by Equation~\eqref{eq:Delta}, and none exceeds
\[
\xi_{\max}\equiv\max\Big\{\log\frac{p_{\mathrm{new}}}{p_{\mathrm{old}}},\ \log\frac{1-p_{\mathrm{new}}}{1-p_{\mathrm{old}}}\Big\}>0.
\]
Since $\fbar^{\mD}_t=q^{\mD}_tp_{\mathrm{new}}+(1-q^{\mD}_t)p_{\mathrm{old}}$,
\begin{equation}
\fbar^{\mD}_t-p_{\mathrm{new}}=(1-q^{\mD}_t)(p_{\mathrm{old}}-p_{\mathrm{new}}).
\label{eq:darwinGapShare}
\end{equation}

(i) By the strong law of large numbers, $z_t/t\to\Delta$ with probability one, so $z_t\to\infty$ and $q^{\mD}_t\to1$. Since $1-q^{\mD}_t=1/(1+e^{z_t})$,
\[
\frac1t\log(1-q^{\mD}_t)=-\frac{z_t}{t}-\frac1t\log\big(1+e^{-z_t}\big)\to-\Delta.
\]
By Equation~\eqref{eq:darwinGapShare}, $\fbar^{\mD}_t\to p_{\mathrm{new}}$, and \Cref{lem:regret}(c) gives
\[
\Rcal_{p_{\mathrm{new}}}(\fbar^{\mD}_t)/(\fbar^{\mD}_t-p_{\mathrm{new}})^2\to1/(2p_{\mathrm{new}}(1-p_{\mathrm{new}})).
\]
Therefore
\[
\frac1t\log\Rcal_{p_{\mathrm{new}}}(\fbar^{\mD}_t)
=\frac2t\log(1-q^{\mD}_t)+\frac2t\log|p_{\mathrm{old}}-p_{\mathrm{new}}|+\frac1t\log\frac{\Rcal_{p_{\mathrm{new}}}(\fbar^{\mD}_t)}{(\fbar^{\mD}_t-p_{\mathrm{new}})^2}
\to-2\Delta,
\]
that is, $\kD=2\Delta$ with probability one.

(ii) Fix $\varepsilon\in(0,1)$, and let
\[
\phi(q)\equiv\Rcal_{p_{\mathrm{new}}}\big(qp_{\mathrm{new}}+(1-q)p_{\mathrm{old}}\big),\qquad q\in[0,1],
\]
be the regret of a Darwinian population in which the adapted lineage has the share $q$. The distance of $qp_{\mathrm{new}}+(1-q)p_{\mathrm{old}}$ from $p_{\mathrm{new}}$ is $(1-q)|p_{\mathrm{old}}-p_{\mathrm{new}}|$, so by \Cref{lem:regret}(a), $\phi$ is continuous and strictly decreasing, from $\phi(0)=\Rcal_{p_{\mathrm{new}}}(p_{\mathrm{old}})=\Delta$ to $\phi(1)=0$. In particular, $\Rcal_{p_{\mathrm{new}}}(\fbar^{\mD}_t)=\phi(q^{\mD}_t)$, and $\phi(q_0)\in(0,\Delta)$. Let $q^*(\varepsilon,q_0)\in(q_0,1)$ be the unique share with
\[
\phi\big(q^*(\varepsilon,q_0)\big)=\varepsilon\,\phi(q_0),
\]
and let $q_\varepsilon\in(0,1)$ be the unique share with $\phi(q_\varepsilon)=\varepsilon\Delta$. Since $\phi$ is strictly decreasing, $\Rcal_{p_{\mathrm{new}}}(\fbar^{\mD}_t)\le\varepsilon\Rcal_{p_{\mathrm{new}}}(\fbar^{\mD}_0)$ if and only if $q^{\mD}_t\ge q^*(\varepsilon,q_0)$, that is, if and only if $\xi_1+\cdots+\xi_t\ge d(\varepsilon,q_0)$, where
\[
d(\varepsilon,q_0)\equiv\log\frac{q^*(\varepsilon,q_0)}{1-q^*(\varepsilon,q_0)}-\log\frac{q_0}{1-q_0}>0
\]
is the distance that the logarithm of the ratio of the two lineages' sizes must travel. Hence $\tau^{\mD}(\varepsilon,q_0)$ is the first passage time of \Cref{lem:passage} with $d=d(\varepsilon,q_0)$ and $\mu=\Delta$, and
\begin{equation}
\frac{d(\varepsilon,q_0)}{\Delta}\le\E\,\tau^{\mD}(\varepsilon,q_0)<\frac{d(\varepsilon,q_0)+\xi_{\max}}{\Delta}.
\label{eq:wald}
\end{equation}

We now bound $d(\varepsilon,q_0)$, which is the sum of two terms:
\[
d(\varepsilon,q_0)=\log\frac{q^*(\varepsilon,q_0)}{q_0}+\log\frac{1-q_0}{1-q^*(\varepsilon,q_0)}.
\]
For the first term, $\phi(q^*(\varepsilon,q_0))=\varepsilon \phi(q_0)<\varepsilon\Delta=\phi(q_\varepsilon)$, so $q^*(\varepsilon,q_0)>q_\varepsilon$; together with $q^*(\varepsilon,q_0)<1$, this gives
\begin{equation}
\log\frac1{q_0}+\log q_\varepsilon<\log\frac{q^*(\varepsilon,q_0)}{q_0}<\log\frac1{q_0}.
\label{eq:firstTerm}
\end{equation}
For the second term, let $v\equiv\min\{p_{\mathrm{old}}(1-p_{\mathrm{old}}),p_{\mathrm{new}}(1-p_{\mathrm{new}})\}$. Every $x$ between $p_{\mathrm{new}}$ and $qp_{\mathrm{new}}+(1-q)p_{\mathrm{old}}$ lies between $p_{\mathrm{old}}$ and $p_{\mathrm{new}}$, where $x(1-x)\ge v$ because $x(1-x)$ is concave. By \Cref{lem:regret}(b), for every $q\in[0,1]$,
\[
2(1-q)^2(p_{\mathrm{new}}-p_{\mathrm{old}})^2\le \phi(q)\le\frac{(1-q)^2(p_{\mathrm{new}}-p_{\mathrm{old}})^2}{2v}.
\]
Applying the upper bound at $q^*(\varepsilon,q_0)$ and the lower bound at $q_0$, and then the lower bound at $q^*(\varepsilon,q_0)$ and the upper bound at $q_0$, the equation $\phi(q^*(\varepsilon,q_0))=\varepsilon \phi(q_0)$ gives
\begin{equation}
4\varepsilon v\le\Big(\frac{1-q^*(\varepsilon,q_0)}{1-q_0}\Big)^2\le\frac{\varepsilon}{4v},
\qquad\text{so}\qquad
\frac12\log\frac{4v}{\varepsilon}\le\log\frac{1-q_0}{1-q^*(\varepsilon,q_0)}\le\frac12\log\frac{1}{4\varepsilon v}.
\label{eq:secondTerm}
\end{equation}

For the first limit in Equation~\eqref{eq:darwinTime}, fix $q_0$. By Equation~\eqref{eq:firstTerm}, the first term of $d(\varepsilon,q_0)$ lies in $(0,\log(1/q_0))$, because $q^*(\varepsilon,q_0)>q_0$. By Equation~\eqref{eq:secondTerm}, the second term differs from $\frac12\log(1/\varepsilon)$ by at most $\frac12\log(1/(4v))$. Hence $d(\varepsilon,q_0)/\log(1/\varepsilon)\to\frac12$ as $\varepsilon\to0$, and dividing Equation~\eqref{eq:wald} by $\log(1/\varepsilon)$ gives $\E\,\tau^{\mD}(\varepsilon,q_0)/\log(1/\varepsilon)\to1/(2\Delta)=1/\kD$.

For the second limit, fix $\varepsilon$. Adding Equations~\eqref{eq:firstTerm} and~\eqref{eq:secondTerm} and using Equation~\eqref{eq:wald},
\[
\frac1\Delta\Big(\log q_\varepsilon+\frac12\log\frac{4v}{\varepsilon}\Big)
<\E\,\tau^{\mD}(\varepsilon,q_0)-\frac{\log(1/q_0)}{\Delta}
<\frac1\Delta\Big(\frac12\log\frac{1}{4\varepsilon v}+\xi_{\max}\Big).
\]
Neither bound depends on $q_0$. Dividing by $\log(1/q_0)$, which tends to infinity as $q_0\to0$, gives $\E\,\tau^{\mD}(\varepsilon,q_0)/\log(1/q_0)\to1/\Delta$.
\end{proof}

\subsubsection{Proofs for Section~\ref{sec:lamarckian}}

\begin{proof}[Proof of \Cref{thm:lam}]
Let $q_0\in[0,1)$. Iterating Equation~\eqref{eq:lamUpdate} gives $f^{\mL}_t-p_{\mathrm{new}}=(1-\lambda)^t(p_{\mathrm{old}}-p_{\mathrm{new}})$, so $f^{\mL}_t$ lies between $p_{\mathrm{old}}$ and $p_{\mathrm{new}}$ and does not depend on the states. The selection step of Mechanism~\ref{mech:L} is Equation~\eqref{eq:shares} with $f_{1,t}=p_{\mathrm{new}}$ and $f_{2,t}=f^{\mL}_t$. \Cref{lem:sandwich} therefore applies with $f_t=f^{\mL}_t$, and, with $K$ from Equation~\eqref{eq:Kdef},
\begin{equation}
\frac{(1-\lambda)^{2t}}{K}\le\frac{\Rcal_{p_{\mathrm{new}}}(\fbar^{\mL}_t)}{\Rcal_{p_{\mathrm{new}}}(\fbar^{\mL}_0)}\le K(1-\lambda)^{2t}
\label{eq:lamSandwich}
\end{equation}
for every $t$ and every history. The constant $K$ depends on neither $q_0$ nor the history.

(i) By Equation~\eqref{eq:lamSandwich}, along every history, $\log\Rcal_{p_{\mathrm{new}}}(\fbar^{\mL}_t)$ differs from $2t\log(1-\lambda)$ by at most $\log K+|\log\Rcal_{p_{\mathrm{new}}}(\fbar^{\mL}_0)|$, which does not depend on $t$. Dividing by $t$ and letting $t\to\infty$ gives $\kL=-2\log(1-\lambda)$.

(ii) Since $(1-\lambda)^{2t}=e^{-\kL t}$, Equation~\eqref{eq:lamSandwich} shows that $\Rcal_{p_{\mathrm{new}}}(\fbar^{\mL}_t)/\Rcal_{p_{\mathrm{new}}}(\fbar^{\mL}_0)>\varepsilon$ whenever $e^{-\kL t}/K>\varepsilon$, and that it is at most $\varepsilon$ whenever $Ke^{-\kL t}\le\varepsilon$. Hence, along every history,
\begin{equation}
\frac{\log(1/(K\varepsilon))}{\kL}\le\tau^{\mL}(\varepsilon,q_0)<\frac{\log(K/\varepsilon)}{\kL}+1.
\label{eq:lamTimeBounds}
\end{equation}
Dividing by $\log(1/\varepsilon)$ and letting $\varepsilon\to0$ shows that $\tau^{\mL}(\varepsilon,q_0)/\log(1/\varepsilon)\to1/\kL$ along every history; since the bounds in Equation~\eqref{eq:lamTimeBounds} do not depend on the history, the same holds for $\E\,\tau^{\mL}(\varepsilon,q_0)/\log(1/\varepsilon)$, which gives the first statement. For the second, the upper bound in Equation~\eqref{eq:lamTimeBounds} does not depend on $q_0$, so $\sup_{q_0\in[0,1)}\E\,\tau^{\mL}(\varepsilon,q_0)\le\log(K/\varepsilon)/\kL+1<\infty$.
\end{proof}

\subsubsection{Proofs for Section~\ref{sec:agentic}}

\begin{proof}[Proof of \Cref{thm:agent}]
Let $q_0\in[0,1)$. The search, evaluation, and learning steps of Mechanism~\ref{mech:A} are Equation~\eqref{eq:searchTarget} with $\hat{p}=p_{\mathrm{new}}$, and \Cref{lem:search}(a) is Equation~\eqref{eq:agentStep}. By \Cref{lem:search}(b), for every sequence of draws, $f^{\mA}_t$ lies between $p_{\mathrm{old}}$ and $p_{\mathrm{new}}$, $|f^{\mA}_t-p_{\mathrm{new}}|=|p_{\mathrm{old}}-p_{\mathrm{new}}|e^{-S_t}$, and $|f^{\mA}_t-p_{\mathrm{new}}|\le(1-\lambda)^t|p_{\mathrm{old}}-p_{\mathrm{new}}|$. The draws do not depend on the states, and the selection step of Mechanism~\ref{mech:A} is Equation~\eqref{eq:shares} with $f_{1,t}=p_{\mathrm{new}}$ and $f_{2,t}=f^{\mA}_t$. \Cref{lem:sandwich} therefore applies with $f_t=f^{\mA}_t$:
\begin{equation}
\frac{e^{-2S_t}}{K}\le\frac{\Rcal_{p_{\mathrm{new}}}(\fbar^{\mA}_t)}{\Rcal_{p_{\mathrm{new}}}(\fbar^{\mA}_0)}\le Ke^{-2S_t}
\label{eq:agentSandwich}
\end{equation}
for every $t$, every history, and every sequence of draws, with $K$ from Equation~\eqref{eq:Kdef}.

(i) By \Cref{lem:search}, $S_t/t\to-\log(1-\lambda)+G_n(\rho)=\kL/2+G_n(\rho)$ with probability one, and by Equation~\eqref{eq:agentSandwich}, $-\frac1t\log\Rcal_{p_{\mathrm{new}}}(\fbar^{\mA}_t)$ differs from $2S_t/t$ by at most $(\log K+|\log\Rcal_{p_{\mathrm{new}}}(\fbar^{\mA}_0)|)/t$. Hence $\kA=\kL+2G_n(\rho)$ with probability one. The variable $-\log(1-\rho U^*_t)$ is nonnegative, so $G_n(\rho)\ge0$, and $G_0(\rho)=0$ because $U^*_t\equiv0$ when $n=0$.

(ii) Each term of $S_t$ is at least $-\log(1-\lambda)=\kL/2>0$, so $S_t$ rises strictly with $t$ and $2S_t\ge\kL t$. For $x\in\mathbb{R}$, let $T(x)\equiv\min\{t\ge0:\ 2S_t\ge x\}$, so that $T(x)=0$ for $x\le0$ and $T(x)\le\lceil x/\kL\rceil$ for $x>0$. By Equation~\eqref{eq:agentSandwich}, the population's regret is at most $\varepsilon\Rcal_{p_{\mathrm{new}}}(\fbar^{\mA}_0)$ once $2S_t\ge\log(K/\varepsilon)$, and it exceeds $\varepsilon\Rcal_{p_{\mathrm{new}}}(\fbar^{\mA}_0)$ while $2S_t<\log(1/(K\varepsilon))$. Since $S_t$ rises with $t$,
\begin{equation}
T\big(\log(1/(K\varepsilon))\big)\le\tau^{\mA}(\varepsilon,q_0)\le T\big(\log(K/\varepsilon)\big)\le\Big\lceil\frac{\log(K/\varepsilon)}{\kL}\Big\rceil
\label{eq:agentTimeBounds}
\end{equation}
for every history and every sequence of draws. The last bound does not depend on $q_0$, which gives the second statement in Equation~\eqref{eq:agentTime}. For the first, $T(x)\to\infty$ as $x\to\infty$, and $2S_{T(x)-1}<x\le2S_{T(x)}$ for $x>0$, so $2S_t/t\to\kA$ gives $T(x)/x\to1/\kA$ with probability one. Since $\log(1/(K\varepsilon))/\log(1/\varepsilon)$ and $\log(K/\varepsilon)/\log(1/\varepsilon)$ tend to one as $\varepsilon\to0$, Equation~\eqref{eq:agentTimeBounds} gives $\tau^{\mA}(\varepsilon,q_0)/\log(1/\varepsilon)\to1/\kA$ with probability one. For $\varepsilon\le\frac12$, the same equation bounds this ratio by $(1+\log K/\log2)/\kL+1/\log2$, so the bounded convergence theorem gives $\E\,\tau^{\mA}(\varepsilon,q_0)/\log(1/\varepsilon)\to1/\kA$.
\end{proof}

\begin{proof}[Proof of \Cref{prop:gain}]
We use the integral form of $G_n(\rho)$ in Equation~\eqref{eq:Gint} of \Cref{lem:Gint}.

(i) For $u\in(0,1)$, the integrand $\rho(1-u^n)/(1-\rho u)$ rises strictly with $n$, because $u^n$ falls. For $n\ge1$, it rises strictly with $\rho$, because $\rho/(1-\rho u)$ does and $1-u^n>0$.

(ii) For $\rho=1$, the integrand is $(1-u^n)/(1-u)=1+u+\cdots+u^{n-1}$, whose integral over $[0,1]$ is $1+\frac12+\cdots+\frac1n$. Since $1/u$ is decreasing, $\int_k^{k+1}du/u\le1/k$ for $k\ge1$ and $1/k\le\int_{k-1}^kdu/u$ for $k\ge2$, so
\[
\log(n+1)\le1+\frac12+\cdots+\frac1n\le1+\log n.
\]
Dividing by $\log n$ and letting $n\to\infty$ gives $G_n(1)/\log n\to1$.

(iii) For $\rho\in(0,1)$, the integrand is below $\rho/(1-\rho u)$ for $u\in(0,1]$, and $\int_0^1\rho/(1-\rho u)\,du=-\log(1-\rho)$, so $G_n(\rho)<-\log(1-\rho)$. As $n\to\infty$, the integrand rises to $\rho/(1-\rho u)$, and the monotone convergence theorem gives $G_n(\rho)\to-\log(1-\rho)$.
\end{proof}

\begin{proof}[Proof of \Cref{prop:Nstar}]
By Equation~\eqref{eq:Gint}, for $n\ge0$,
\begin{equation}
G_{n+1}(\rho)-G_n(\rho)=\int_0^1\frac{\rho\,u^n(1-u)}{1-\rho u}\,du,
\label{eq:Gincrement}
\end{equation}
which is positive and strictly decreasing in $n$.

(i) By Equation~\eqref{eq:longRunCost}, cross-multiplying shows that $C(n+1)$ is below, equal to, or above $C(n)$ according as
\[
\psi(n)\equiv2(1+cn)\big(G_{n+1}(\rho)-G_n(\rho)\big)-c\big(\kL+2G_n(\rho)\big)
\]
is positive, zero, or negative. The function $\psi$ is strictly decreasing:
\[
\psi(n+1)-\psi(n)=2\big(1+c(n+1)\big)\Big[\big(G_{n+2}(\rho)-G_{n+1}(\rho)\big)-\big(G_{n+1}(\rho)-G_n(\rho)\big)\Big]<0.
\]
It is also negative for large $n$. Since $\rho(1-u)/(1-\rho u)\le1$, Equation~\eqref{eq:Gincrement} gives $G_{n+1}(\rho)-G_n(\rho)\le1/(n+1)$. For $\rho=1$, the first term of $\psi(n)$ is therefore at most $2(1+cn)/(n+1)\le2\max\{1,c\}$, while the second, $c(\kL+2(1+\frac12+\cdots+\frac1n))$ by \Cref{prop:gain}(ii), grows without bound. For $\rho\in(0,1)$, $\rho(1-u)/(1-\rho u)\le\rho(1-u)/(1-\rho)$, so the first term is at most $2(1+cn)\rho/((1-\rho)(n+1)(n+2))$, which tends to zero, while the second is at least $c\kL>0$. Hence there is an integer $\bar n\ge0$ such that $C(n+1)<C(n)$ for $n<\bar n$, $C(\bar n+1)\ge C(\bar n)$, and $C(n+1)>C(n)$ for $n>\bar n$: the long-run cost first decreases and then increases in $n$. Its minimizers are $\bar n$, and also $\bar n+1$ when $C(\bar n+1)=C(\bar n)$.

(ii) For $n'>n\ge0$,
\[
\frac{C(n')}{C(n)}=\frac{1+cn'}{1+cn}\cdot\frac{\kL+2G_n(\rho)}{\kL+2G_{n'}(\rho)}.
\]
The first factor rises strictly with $c$, because its derivative in $c$ has the sign of $n'-n$, and does not depend on $\lambda$ or $\rho$. The second factor does not depend on $c$. It rises strictly with $\kL=-2\log(1-\lambda)$, and so with $\lambda$, because $G_{n'}(\rho)>G_n(\rho)$ by \Cref{prop:gain}(i). It falls strictly with $\rho$. For $n=0$, this follows from \Cref{prop:gain}(i), because $G_0(\rho)=0$. For $n\ge1$, expanding $1/(1-\rho u)=\sum_{k\ge0}(\rho u)^k$ in Equation~\eqref{eq:Gint} and integrating term by term gives, for $\rho\in(0,1)$,
\[
\kL+2G_n(\rho)=\sum_{k\ge0}\nu_k(n)\rho^k,
\qquad
\nu_0(n)=\kL,\quad\nu_k(n)=\frac{2n}{k(n+k)}\ \ (k\ge1).
\]
For $n'>n\ge1$, the ratio $\nu_k(n')/\nu_k(n)$ equals one for $k=0$ and $n'(n+k)/(n(n'+k))$ for $k\ge1$, which exceeds one and rises with $k$. \Cref{lem:series}, with $a_k=\nu_k(n')$ and $b_k=\nu_k(n)$, shows that $(\kL+2G_{n'}(\rho))/(\kL+2G_n(\rho))$ rises strictly with $\rho$ on $(0,1)$; since $G_n$ and $G_{n'}$ are continuous on $(0,1]$ by Equation~\eqref{eq:Gint}, it rises strictly on $(0,1]$.

Consider a change in one parameter that lowers $C(n')/C(n)$ for every pair $n'>n$: a rise in $\rho$, or a fall in $\lambda$ or in $c$. Let $n_1$ minimize $C$ before the change and $n_2$ after it, and suppose that $n_2<n_1$. Before the change, $C(n_1)/C(n_2)\le1$. After the change, $C(n_1)/C(n_2)<1$, which contradicts the optimality of $n_2$. Hence every minimizer after the change is at least as large as every minimizer before it, so $n^*(\rho,\lambda,c)$ is nondecreasing in $\rho$ and nonincreasing in $\lambda$ and in $c$.
\end{proof}

\subsubsection{Proofs for Section~\ref{sec:combined}}

\begin{proof}[Proof of \Cref{thm:hierarchy}]
(i) Let $q_0\in(0,1)$. By \Cref{thm:darwin}(i), $\kD=2\Delta$ with probability one. By \Cref{thm:lam}(i), $\kL=-2\log(1-\lambda)$, and Equation~\eqref{eq:speedCondition} is equivalent to $-\log(1-\lambda)>\Delta$, so $\kL>\kD$. By \Cref{thm:agent}(i), $\kA=\kL+2G_n(\rho)$ with probability one, and $G_n(\rho)>G_0(\rho)=0$ for $n\ge1$ and $\rho\in(0,1]$ by \Cref{prop:gain}(i), so $\kA>\kL$.

(ii) Let $q_0\in(0,1)$. By part (ii) of \Cref{thm:darwin,thm:lam,thm:agent}, the ratios $\E\,\tau^{\mD}(\varepsilon,q_0)/\log(1/\varepsilon)$, $\E\,\tau^{\mL}(\varepsilon,q_0)/\log(1/\varepsilon)$, and $\E\,\tau^{\mA}(\varepsilon,q_0)/\log(1/\varepsilon)$ tend to $1/\kD$, $1/\kL$, and $1/\kA$ as $\varepsilon\to0$. By part (i), $1/\kA<1/\kL<1/\kD$. Let $\delta$ be half of the smaller of the gaps $1/\kL-1/\kA$ and $1/\kD-1/\kL$. There is $\bar\varepsilon\in(0,1)$ such that, for every $\varepsilon\in(0,\bar\varepsilon)$, each ratio lies within $\delta$ of its limit, and multiplying by $\log(1/\varepsilon)>0$ gives Equation~\eqref{eq:timeRanking}.
\end{proof}

\subsubsection{Proofs for Section~\ref{sec:safety}}

\begin{proof}[Proof of \Cref{thm:pressure}]
Let $\gamma\equiv\min\{p_{\mathrm{old}},1-p_{\mathrm{old}},p_{\mathrm{new}},1-p_{\mathrm{new}},\tilde{p},1-\tilde{p}\}>0$. The search, evaluation, and learning steps of Mechanism~\ref{mech:F} are Equation~\eqref{eq:searchTarget} with $\hat{p}=\tilde{p}\ne p_{\mathrm{old}}$. By \Cref{lem:search}, for every sequence of draws, $f^{\mF}_t$ lies between $p_{\mathrm{old}}$ and $\tilde{p}$, so in $[\gamma,1-\gamma]$, and
\begin{equation}
|f^{\mF}_t-\tilde{p}|=|p_{\mathrm{old}}-\tilde{p}|\,e^{-S_t}\le(1-\lambda)^t|p_{\mathrm{old}}-\tilde{p}|,
\label{eq:flawedGap}
\end{equation}
and $S_t/t\to\kL/2+G_n(\rho)$ with probability one.

(i) Every $x$ between $f^{\mF}_t$ and $\tilde{p}$ lies in $[\gamma,1-\gamma]$, where $x(1-x)\ge\gamma(1-\gamma)$. By \Cref{lem:regret}(b), $\Rcal_{\tilde{p}}(f^{\mF}_t)$ lies between $2(f^{\mF}_t-\tilde{p})^2$ and $(f^{\mF}_t-\tilde{p})^2/(2\gamma(1-\gamma))$. By Equation~\eqref{eq:flawedGap}, $\log\Rcal_{\tilde{p}}(f^{\mF}_t)$ therefore differs from $2\log|p_{\mathrm{old}}-\tilde{p}|-2S_t$ by at most $\log(1/(2\gamma(1-\gamma)))$, so $-\frac1t\log\Rcal_{\tilde{p}}(f^{\mF}_t)\to\kL+2G_n(\rho)$ with probability one.

(ii) By Equation~\eqref{eq:flawedGap}, $f^{\mF}_t\to\tilde{p}$ for every sequence of draws. The function $\Rcal_{p_{\mathrm{new}}}$ is continuous on $(0,1)$, so $\Rcal_{p_{\mathrm{new}}}(f^{\mF}_t)\to\Rcal_{p_{\mathrm{new}}}(\tilde{p})$, which is positive by \Cref{prop:kl} because $\tilde{p}\ne p_{\mathrm{new}}$.

(iii) The selection step of Mechanism~\ref{mech:F} is Equation~\eqref{eq:shares} with $p=p_{\mathrm{new}}$, $f_{1,t}=p_{\mathrm{new}}$, and $f_{2,t}=f^{\mF}_t$. The behaviors $f^{\mF}_t$ do not depend on the states, $p_{\mathrm{new}}$ and every $f^{\mF}_t$ lie in $[\gamma,1-\gamma]$, and $\sum_{t\ge0}|f^{\mF}_t-\tilde{p}|\le|p_{\mathrm{old}}-\tilde{p}|/\lambda$ by Equation~\eqref{eq:flawedGap}. By \Cref{lem:twoLineages}(b), applied for every sequence of draws, $z_t\equiv\log(q^{\mF}_t/(1-q^{\mF}_t))$ satisfies $z_t/t\to\Rcal_{p_{\mathrm{new}}}(\tilde{p})>0$ with probability one. As in the proof of \Cref{thm:darwin}(i), $q^{\mF}_t\to1$ and $\frac1t\log(1-q^{\mF}_t)\to-\Rcal_{p_{\mathrm{new}}}(\tilde{p})$. Since
\[
\fbar^{\mF}_t-p_{\mathrm{new}}=(1-q^{\mF}_t)\big(f^{\mF}_t-p_{\mathrm{new}}\big)
\]
and $f^{\mF}_t-p_{\mathrm{new}}\to\tilde{p}-p_{\mathrm{new}}\ne0$, it follows that $\fbar^{\mF}_t\to p_{\mathrm{new}}$ and $\frac1t\log|\fbar^{\mF}_t-p_{\mathrm{new}}|\to-\Rcal_{p_{\mathrm{new}}}(\tilde{p})$. By \Cref{lem:regret}(c), $\log\Rcal_{p_{\mathrm{new}}}(\fbar^{\mF}_t)-2\log|\fbar^{\mF}_t-p_{\mathrm{new}}|$ converges to $-\log(2p_{\mathrm{new}}(1-p_{\mathrm{new}}))$, so $-\frac1t\log\Rcal_{p_{\mathrm{new}}}(\fbar^{\mF}_t)\to2\Rcal_{p_{\mathrm{new}}}(\tilde{p})$ with probability one.
\end{proof}

\clearpage
\singlespacing
\bibliographystyle{jf3}
\bibliography{references}

\end{document}